\documentclass[letterpaper, 10 pt, conference]{ieeeconf}  % Comment this line out if you need a4paper

\IEEEoverridecommandlockouts                              % This command is only needed if 
\usepackage{graphicx}
\usepackage[table]{xcolor}
\usepackage{adjustbox}
\usepackage{booktabs}

\title{\LARGE \bf
Preliminary analysis of RGB-NIR Image Registration techniques for off-road forestry environments
}

\author{Pankaj Deoli$^{1}$ and Karthik Ranganath$^{2}$ and Karsten Berns$^{1}$% <-this % stops a space
\thanks{$^{1}$ Pankaj Deoli and Karsten Berns are associated with Robotics Research Laboratory, University of Kaiserslautern-Landau (RPTU, Kaiserslautern, Germany)
        {\tt\small {pankaj.deoli, karsten.berns}@cs.rptu.de}}%
\thanks{$^{2}$ The work done by Karthik Ranganath is associated with their Master Thesis with the lab. 
        {\tt\small krangana@rptu.de}}%
}

\begin{document}

\maketitle
\thispagestyle{empty}
\pagestyle{empty}

%%%%%%%%%%%%%%%%%%%%%%%%%%%%%%%%%%%%%%%%%%%%%%%%%%%%%%%%%%%%%%%%%%%%%%%%%%%%%%%%
\begin{abstract}

RGB-NIR image registration plays an important role in sensor-fusion, image enhancement and off-road autonomy. In this work, we evaluate both classical and Deep Learning (DL) based image registration techniques to access their suitability for off-road forestry applications. NeMAR \cite{c12}, trained under 6 different configurations, demonstrates partial success however, its GAN loss instability suggests challenges in preserving geomertic consistency. MURF \cite{c16}, when tested on off-road forestry data shows promising large scale feature alignment during shared information extraction but struggles with fine details in dense vegetation. Even though this is just a preliminary evaluation, our study necessitates further refinements for robust, multi-scale registartion for off-road forest applications.   

\end{abstract}

%%%%%%%%%%%%%%%%%%%%%%%%%%%%%%%%%%%%%%%%%%%%%%%%%%%%%%%%%%%%%%%%%%%%%%%%%%%%%%%%
\section{INTRODUCTION}

Precise image registration is a critical task in robotics and computer vision, enabling concepts like sensor fusion, environment mapping, and localization. Although traditional image registration methods have shown success in well-structured, man-made environments, they often struggle in forests or off-road settings. This challenge becomes even more pronounced if the spectral bands are different (for e.g. RGB and Infrared). A significant amount of spectral discrepancy, together with dense vegetation, irregular illumination, and repetitive textures, leads to unstable feature matches and high alignment errors. As a result, image pairs captured mere seconds-apart, can exhibit stark differences in contrast, color distribution, brightness, thereby making it more difficult for conventional registration algorithms to find consistent correspondences. \\
The challenge is further profound with RGB-NIR image registration, where spectral discrepancies further complicate feature correspondence. While NIR can enhance foliage discovery and improve robustness to certain lighting conditions, it introduces inconsistencies in texture and brightness compared to RGB. Classical feature descriptors fails to bridge this gap and therefore, in this work, we evaluate both the traditional as well as DL models for RGB-NIR image registration.

\section{Related Works}
Feature-based methods (for image registration) typically depend on detecting and describing salient keypoints that stay robust under transformations such as scale changes, rotations, etc. Well-known algorithms like SIFT \cite{c1}, SURF \cite{c2}, ORB \cite{c3} and KAZE \cite{c4} compute distinctive descriptors around each detected key-point, then match the descriptors between 2 images (as shown in table \ref{tab:traditional_methods}). However, in-case of multi-spectral registration, these descriptors often struggle due to large disparities in intensity and texture. Outlier rejection methods such as RANSAC (Random Sample Consensus) \cite{c5}, although being adept at segregating mis-matched key-points, still relies on having a sufficient pool of correct correspondences to converge on a valid solution. Intensity-based method targets registration by directly optimizing a similarity metric across the entire image, thereby diverting the need for explicit key-point extraction.\\
The evolution from these classical approaches to advanced DL approaches has brought remarkable opportunities for addressing such challenges. Early works such as \cite{c6} laid the foundation for multi-modal registration in natural images, while \cite{c7} explored low-light super-resolution through multi-frame RGB/NIR imaging. Later, \cite{c8} introduced SMRD (Shearlet-based Modality Robust Descriptor) - a local descriptor developed for cross-spectral registration to improve correspondences matching. In 2022, studies like \cite{c9} with RFNet, along-with some works focused on plant phenotyping (\cite{c10}) and agricultural monitoring (\cite{c11}), furthered the field, though their applications remain distinct from complex, resolution-diverse conditions as present in forests. in 2020, \cite{c12} introduced an unsupervised method for multi-modal registration (NeMAR) that bypasses the challenges of developing cross-modality similarity measures. However, its application to forestry environments present challenges that require further domain-specific refinements and robust multi-scale fusion strategies.
In 2023, approaches such as DarkVisionNet (\cite{c13}) for low-light fusion, long-range imaging using multi-spectral fusion (\cite{c14}) and attention-based fusion methods such as YOLACTFusion (\cite{c15}), MURF (Mutually Reinforcing Multi-Modal Image Registration and Fusion \cite{c16}), further progressed multi-modal registration by integrating mutually reinforcing registration and fusion processes. However, these techniques generally target urban or agricultural environments with more uniform textures and controlled resolutions than those typical in forests environments. Recent work, however, has shown a shift towards transformer-based and benchmark-driven methods. A comprehensive survey in \cite{c17} synthesized progress in the field, while methods like SGFormer with a new benchmark was introduced in \cite{c18} for RGB-NIR image registration. Cross-modal transformer strategies were further explored in \cite{c19} and the applications were extended to robotic vision \cite{c20} and broader comparative surveys \cite{c21}. \\
Despite these advancements, the challenges of image registration in forests environments - characterized with non-uniform textures, occlusions and significant resolution disparities are not fully addressed by existing methods. Our preliminary experiments which included NeMAR \cite{c12} did not yield satisfactory results along-with early stage evaluations using MURF \cite{c16}, highlight the need for further domain-specific refinements and robust multi-scale fusion strategies to handle the complexity.

\section{Experimental Approaches}
The data, as previously mentioned in \cite{c22}, was further extended to $\sim$5400 RGB images along-with their respective multi-spectral spectrum such as RED, REG (Red-Edge), NIR (Near-Infrared) and GREEN. 

\subsection{Classical approaches}
The results presented in table \ref{tab:image_registration_methods} showcases the performance of three classical image registration techniques i.e. Keypoint matching, Histogram matching and template matching. Upon visual inspection, none of the methods produced satisfactory alignment between the two modalities. Key-point matching failed to find sufficient reliable matches due to significant spectral differences, leading to misaligned output. Histogram matching does not account for geometric distortions or structural differences and therefore did not contribute to meaningful result. Template matching (being sensitive to contrast variations) also did not generalize well for off-road forestry conditions. Apart from this, problems such a scale dependency (RGB, NIR resolutions being different), lack of robust adaptation also hampered Corner detection (Shi-Tomasi) for our application at hand. Additional image registration methods such as Mutual Information and Fourier based image registration also yielded unsatisfactory results due to a high variation in spectra.

\subsection{Deep learning based approaches}
For evaluation of DL approaches to our task at hand, we evaluated NeMAR \cite{c12} and MURF \cite{c16} (only testing for pre-trained model). NeMAR was trained with 6 different configurations as mentioned in table \ref{tab:training_configs}, the results of which are further showcased in table \ref{tab:images}. The approach demonstrates partial success in RGB-NIR alignment, as \textbf{\textit{L1-loss convergence depicts effective pixel-wise transformations}}. However, \textbf{\textit{unstable GAN loss}} across most configurations indicates a challenge in preserving fine-grained structures, thereby leading to a potential misalignment of tree crowns and terrain features.  While higher resolution training (i.e. 960 x 960) shows promise, it requires lower learning rates and significantly longer training times to avoid distorted or blank outputs. Additionally, the \textbf{\textit{model struggles with generalization}}, as training variations in batch size and learning rate impact the stability, thereby making it sensitive to occlusions, seasonal variations and lighting changes. The use of \textbf{\textit{LSGAN (Config 4) improves stability}} but geometry preservation remains inconsistent. \\
From a computation pov, higher batch sizes (Config 5) improves efficiency but risks model collapse, while lower batch sizes (Config 3, 4) better preserve fine details but slows-down training. These trade-offs highlight important suggestions that adaptive learning rate scheduling and progressive resolution scaling are important for forestry-applications. \\
Preliminary testing of MURF for off-road environments highlighted that although the model works well (during shared information extraction) for large-scale features, it loses small scale details like fine branches and subtle foliage variations. During multi-scale coarse registration, based on the rigid structures, the models perform good . however, the high spectral differences between RGB-NIR causes mis-matches in shadowed regions. Finally, during fine registration and fusion, the model manages to align large structures, but struggles with fine details. \\

\begin{table}[htbp!]
    \centering
    \renewcommand{\arraystretch}{1.2}
    \setlength{\tabcolsep}{2pt} % Adjust column spacing for better layout

    \caption{Training Configurations for NeMAR \cite{c12}}
    \label{tab:training_configs}

    \begin{tabular}{cccccc}
        \toprule
        \textbf{Configs} & \textbf{Batch Size} & \textbf{Learning Rate} & \textbf{Image Size} & \textbf{Epochs} & \textbf{GAN Mode} \\
        \midrule
        Config 1 & 32 & 0.0002 & 286p & 100 & Vanilla \\
        Config 2 & 32 & 0.0002 & 600p & 100 & Vanilla \\
        Config 3 & 12 & 0.0001 & 600p & 200 & Vanilla \\
        Config 4 & 12 & 0.0001 & 600p & 200 & LSGAN \\
        Config 5 & 64 & 0.0001 & 600p & 100 & Vanilla \\
        Config 6 & 12 & 0.00005 & 960p & 50 & Vanilla \\
        \bottomrule
    \end{tabular}
\end{table}

\begin{table}[htbp!]
    \centering
    \renewcommand{\arraystretch}{1.3} % Adjust row spacing
    \setlength{\tabcolsep}{2pt} % Reduce column spacing for better compactness

    \caption{Results of different training configurations on NeMAR \cite{c12}}
    \label{tab:images}

    \resizebox{\columnwidth}{!}{ % Resize table to fit within the column width
    \begin{tabular}{m{0.18\textwidth} m{0.18\textwidth} m{0.22\textwidth}} 
        \toprule
        \centering \textbf{L1 Loss} & 
        \centering \textbf{GAN Loss} & 
        \centering \textbf{Discriminator score}  \tabularnewline
        \midrule
        % First Image Pair
        \centering \includegraphics[width=0.16\textwidth, height=0.16\textwidth]{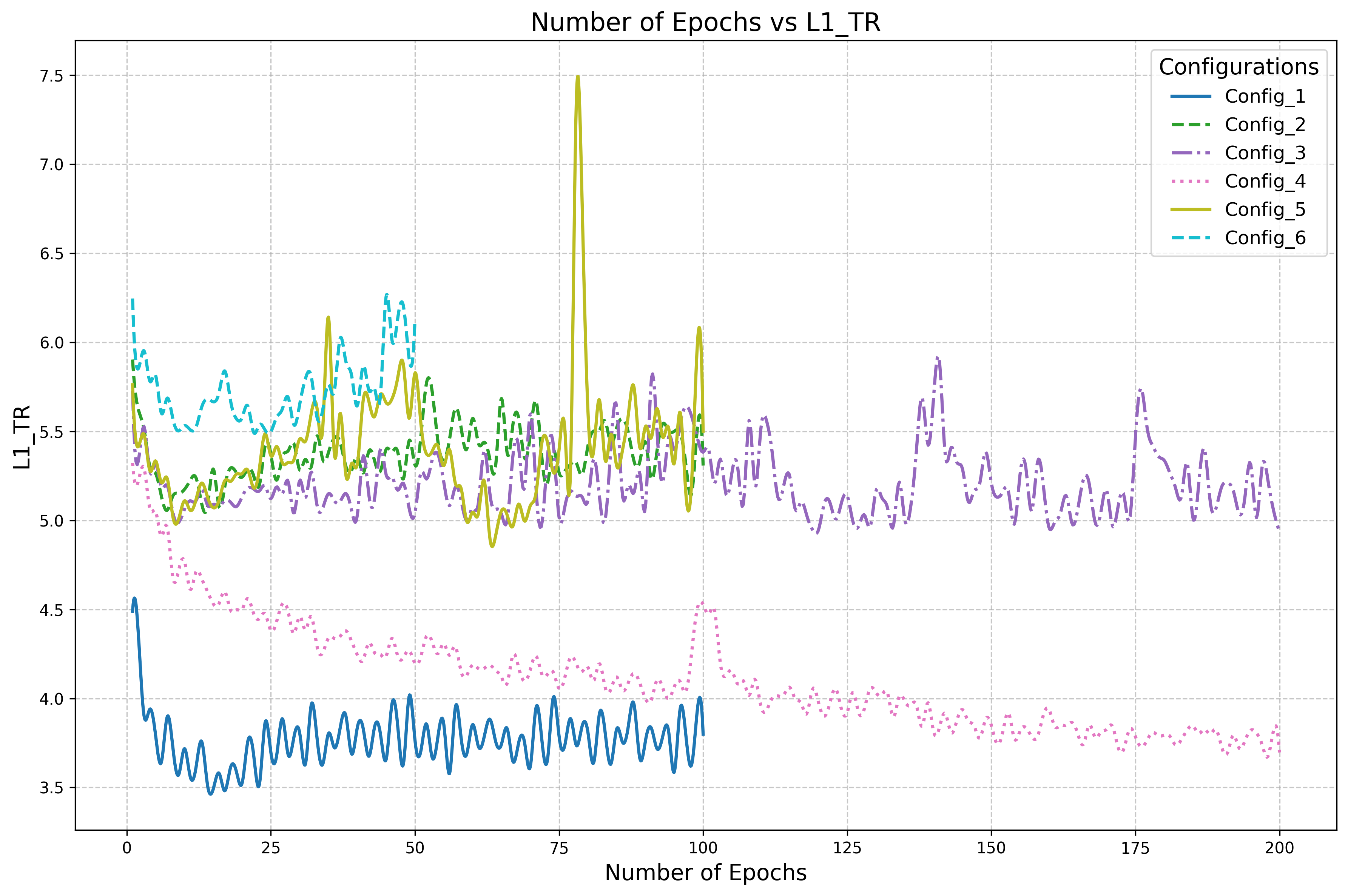} & 
        \centering \includegraphics[width=0.16\textwidth, height=0.16\textwidth]{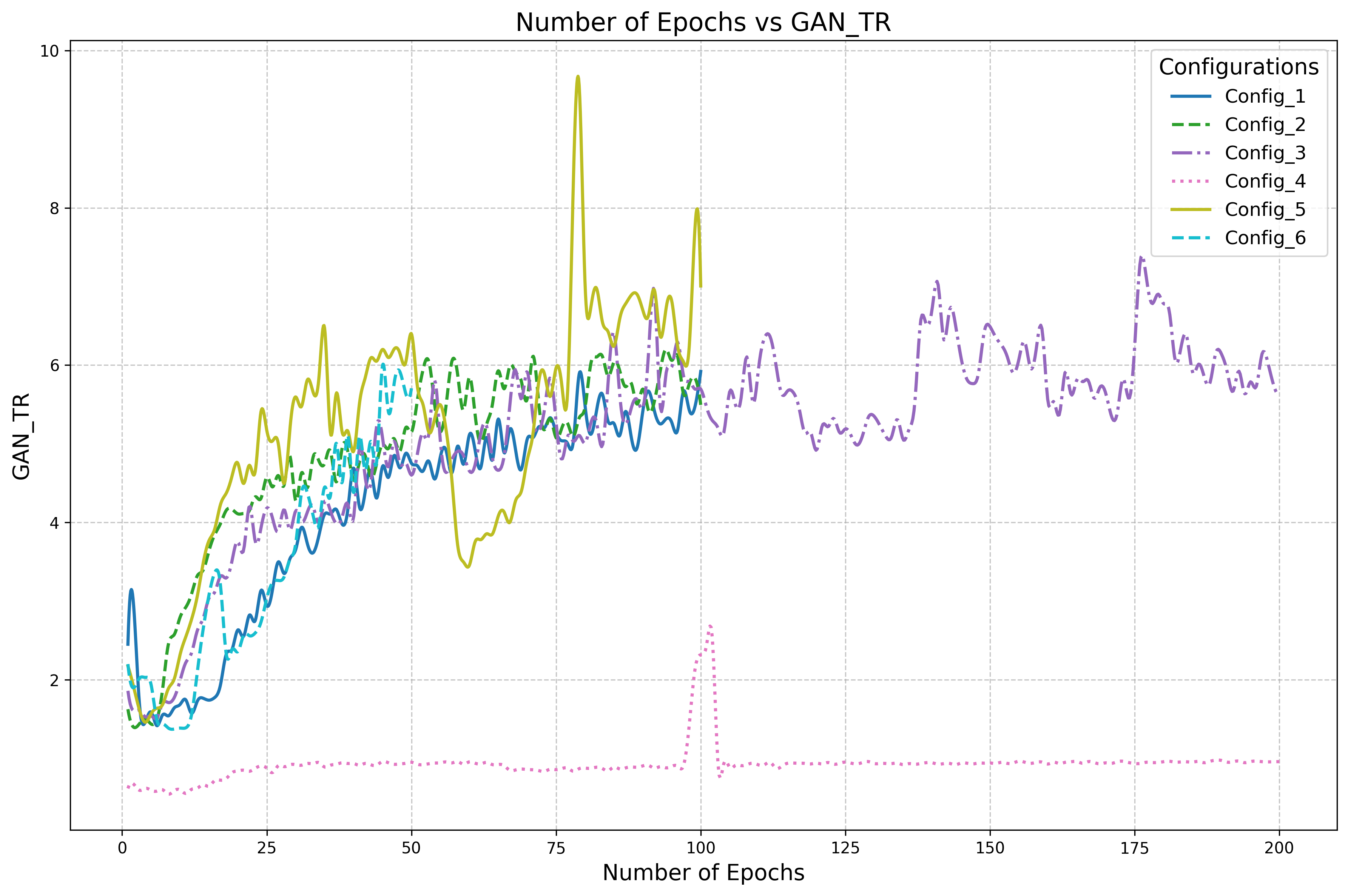} & 
        \centering \includegraphics[width=0.22\textwidth, height=0.16\textwidth]{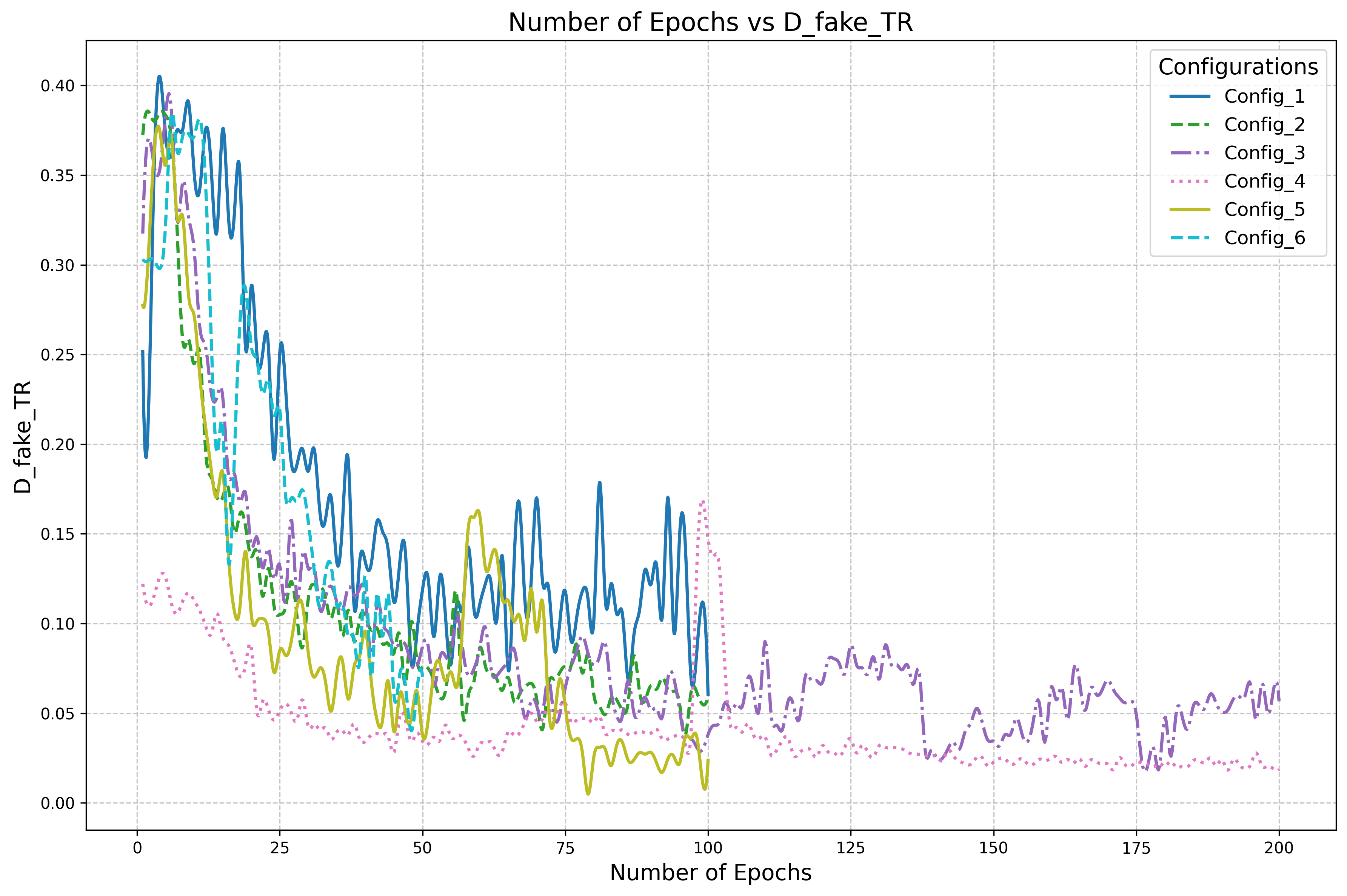}  \tabularnewline
        \midrule
        % Second Image Pair
        \centering \includegraphics[width=0.16\textwidth, height=0.16\textwidth]{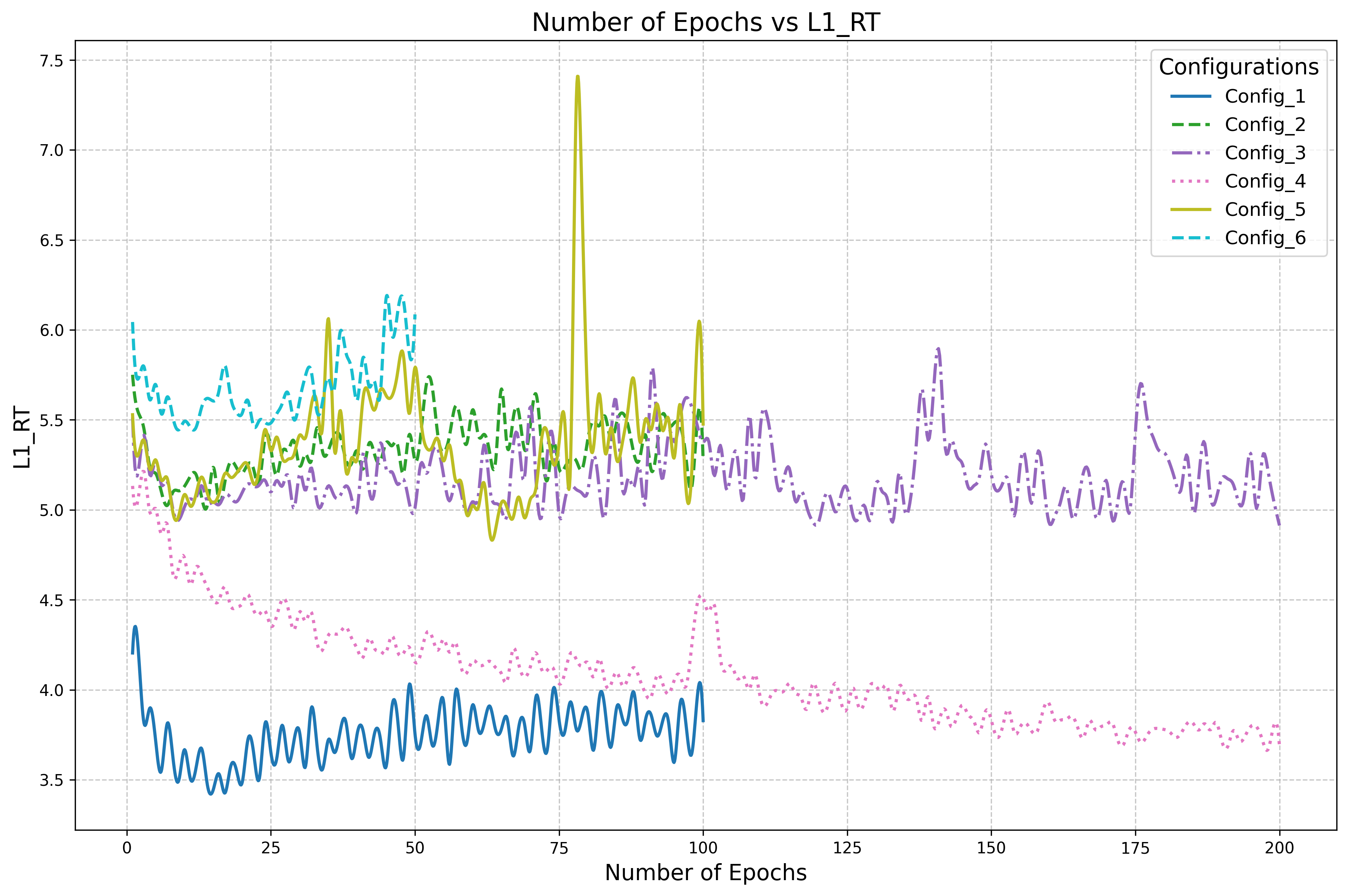} & 
        \centering \includegraphics[width=0.16\textwidth, height=0.16\textwidth]{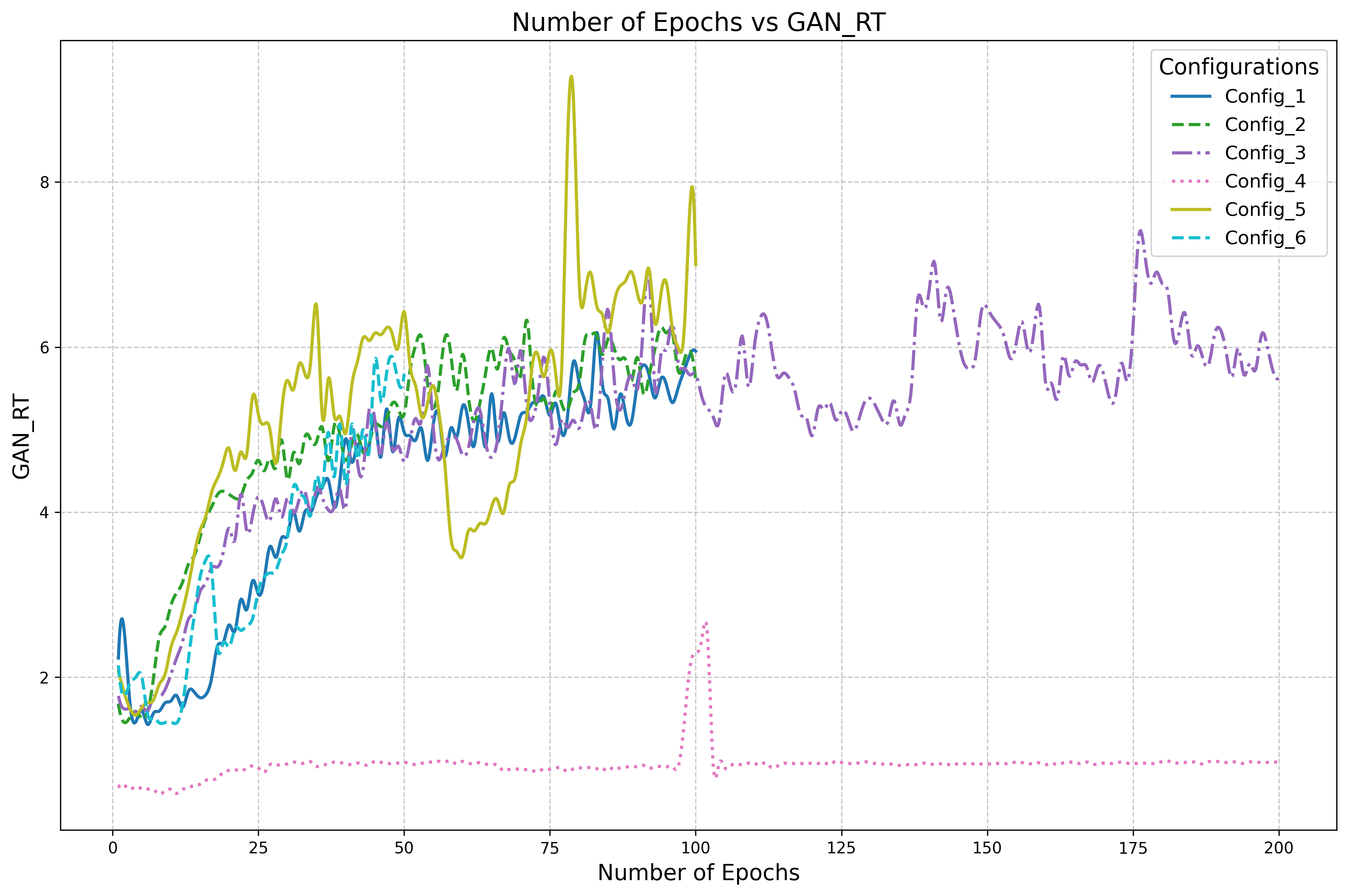} & 
        \centering \includegraphics[width=0.22\textwidth, height=0.16\textwidth]{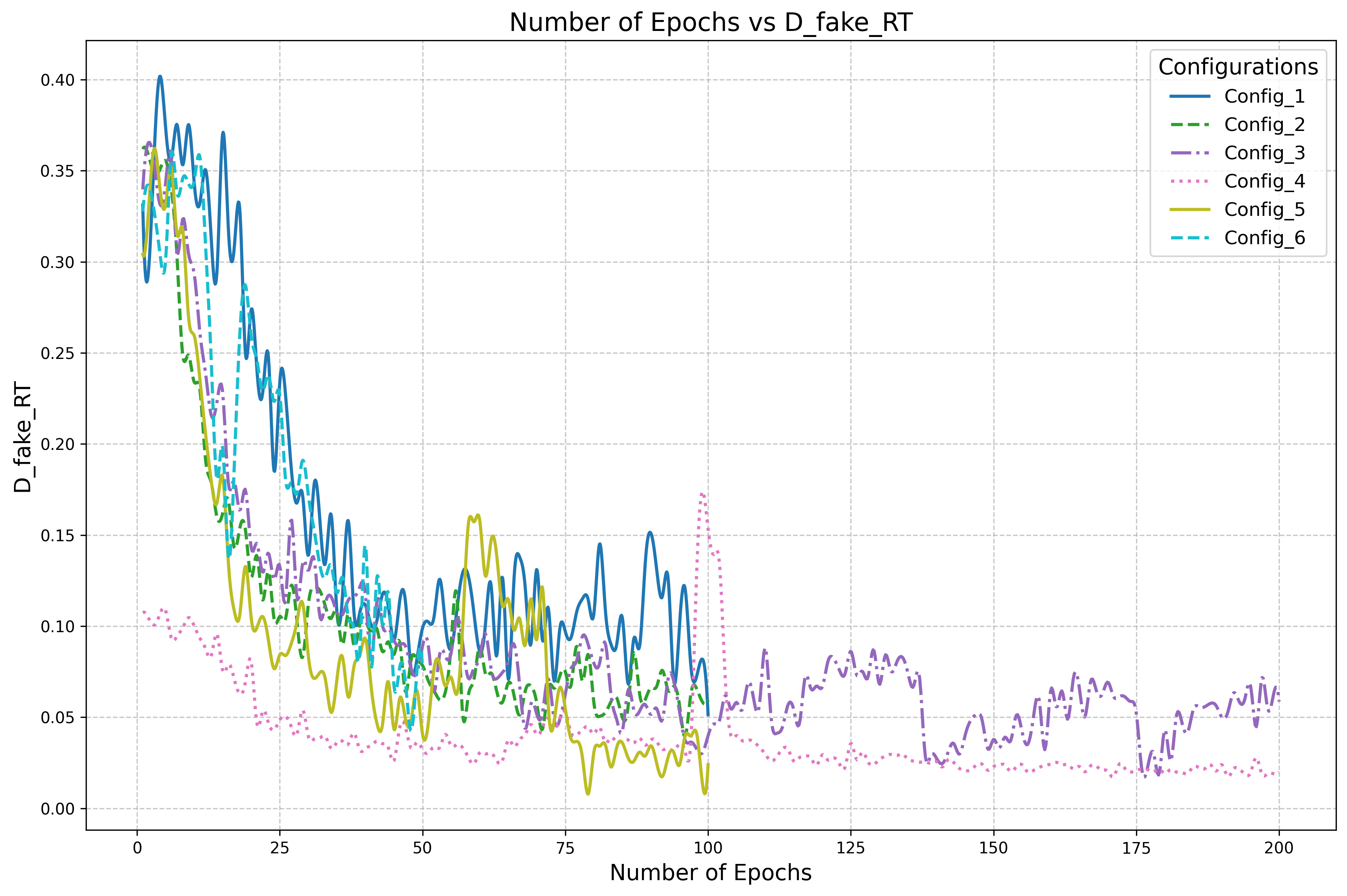}  \tabularnewline
        \bottomrule
    \end{tabular}
    } % End resizebox

\end{table}

\section{CONCLUSIONS}

Our evaluation of RGB-NIR image registration techniques for off-road forestry environments reveals that both classical and deep learning-based methods face unique challenges due to spectral mismatches, dense vegetation, and varying lighting conditions. Even though the DL techniques gives partial success, these findings emphasize the need for domain-specific enhancements. Future work will explore transformer-based architectures and attention-driven fusion techniques to enhance robustness in multi-modal image alignment.

\addtolength{\textheight}{0cm}   % This command serves to balance the column lengths

%%%%%%%%%%%%%%%%%%%%%%%%%%%%%%%%%%%%%%%%%%%%%%%%%%%%%%%%%%%%%%%%%%%%%%%%%%%%%%%%

%%%%%%%%%%%%%%%%%%%%%%%%%%%%%%%%%%%%%%%%%%%%%%%%%%%%%%%%%%%%%%%%%%%%%%%%%%%%%%%%

%%%%%%%%%%%%%%%%%%%%%%%%%%%%%%%%%%%%%%%%%%%%%%%%%%%%%%%%%%%%%%%%%%%%%%%%%%%%%%%%

%%%%%%%%%%%%%%%%%%%%%%%%%%%%%%%%%%%%%%%%%%%%%%%%%%%%%%%%%%%%%%%%%%%%%%%%%%%%%%%%

\section*{Appendix}

\begin{table}[htbp]
    \centering
    \renewcommand{\arraystretch}{1.3} % Adjust row spacing
    \setlength{\tabcolsep}{3pt} % Adjust column spacing

    \caption{Comparison of Different Image Registration Methods}
    \label{tab:image_registration_methods}

    \resizebox{\columnwidth}{!}{ % Resize to fit column width
    \begin{tabular}{m{0.18\textwidth} m{0.18\textwidth} m{0.36\textwidth} m{0.18\textwidth} m{0.18\textwidth}} % Keypoint Matching is wider
        \toprule
        \centering \textbf{RGB Image} & 
        \centering \textbf{NIR Image} & 
        \centering \textbf{Keypoint Matching (Increased Height)} & 
        \centering \textbf{Histogram Matching} & 
        \centering \textbf{Template Matching} \tabularnewline
        \midrule
        \centering \includegraphics[angle=-180,origin=c, width=0.16\textwidth]{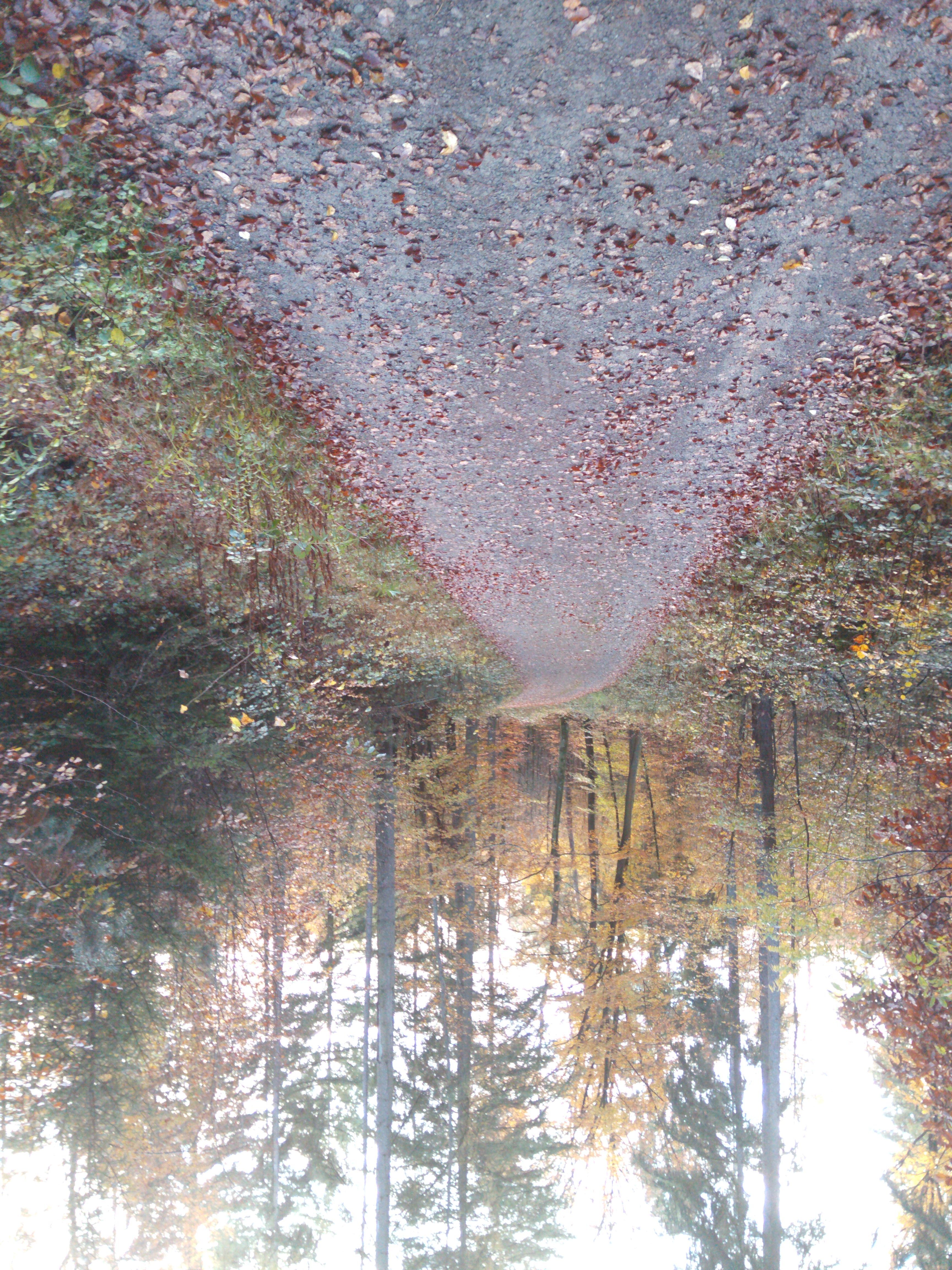} & 
        \centering \includegraphics[width=0.16\textwidth]{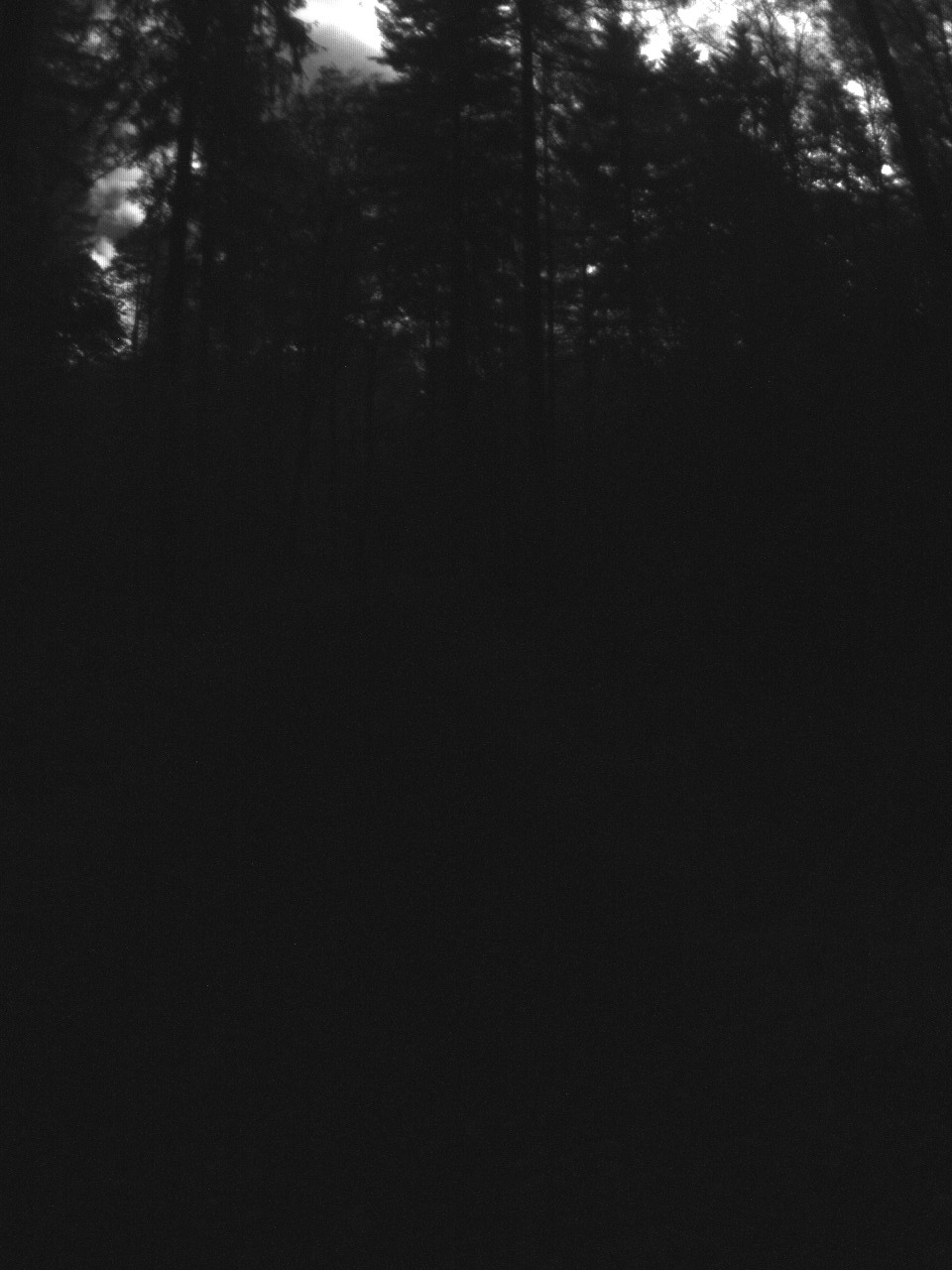} & 
        \centering \includegraphics[width=0.32\textwidth, height=0.22\textwidth]{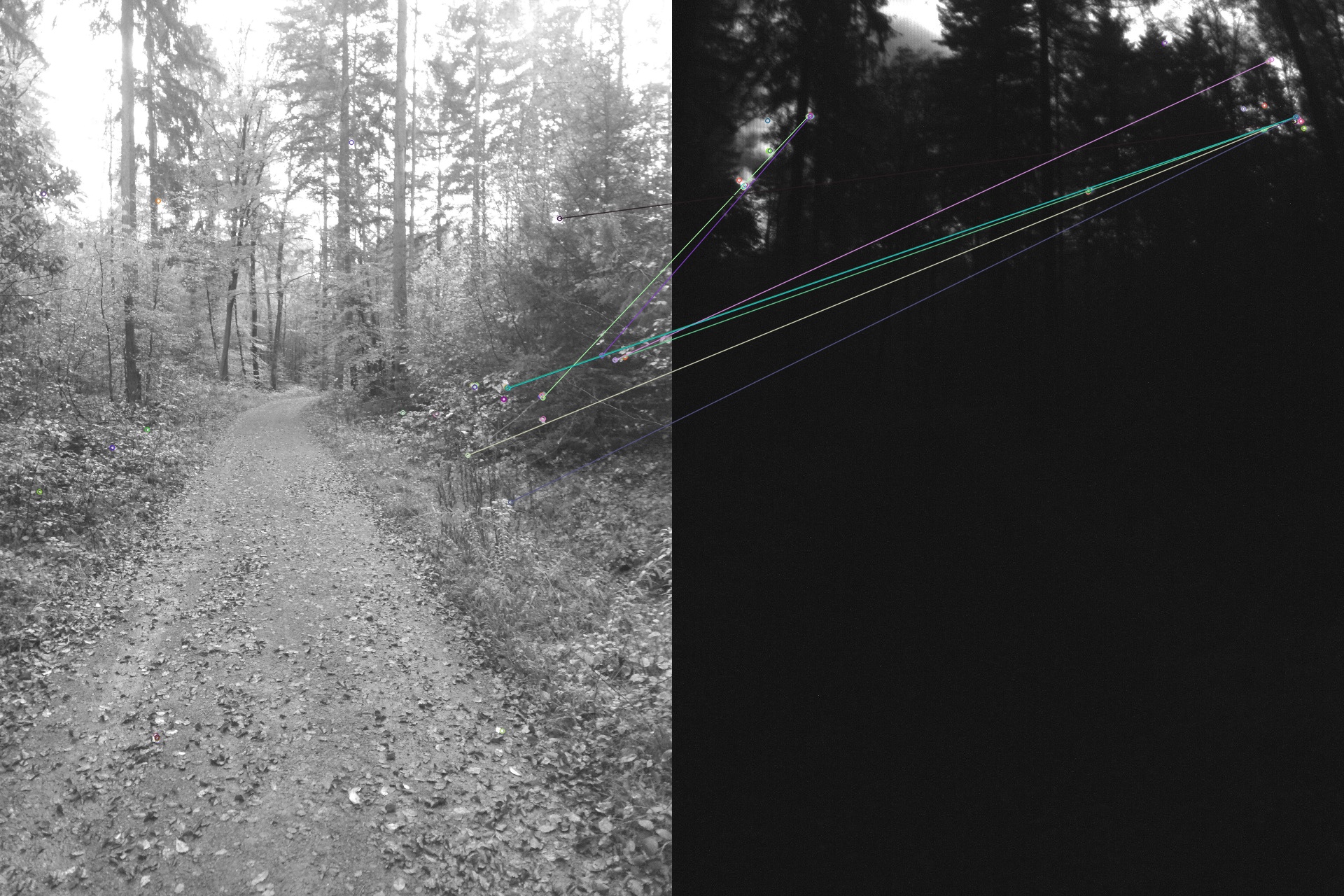} & 
        \centering \includegraphics[width=0.16\textwidth]{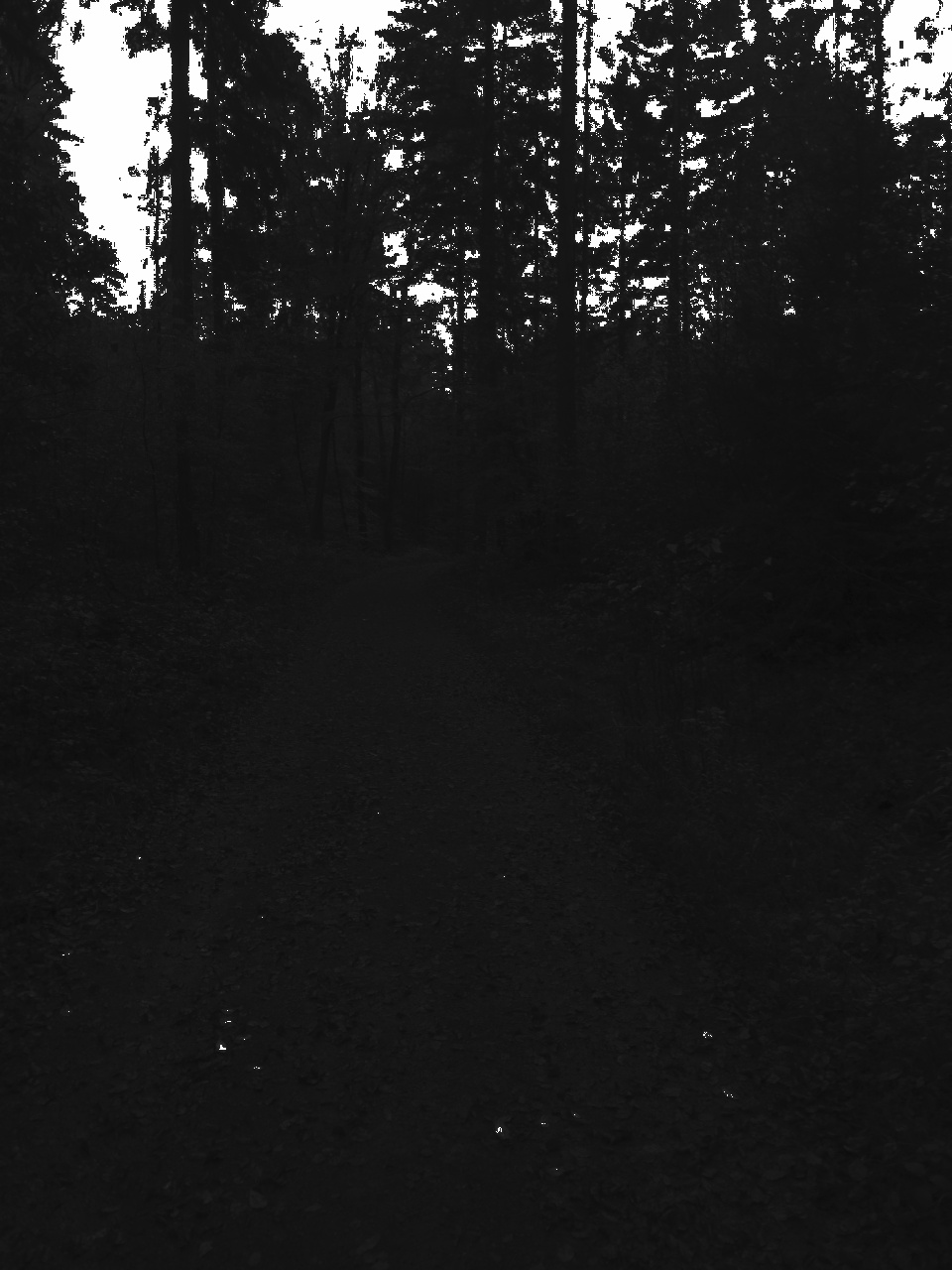} & 
        \centering \includegraphics[width=0.16\textwidth]{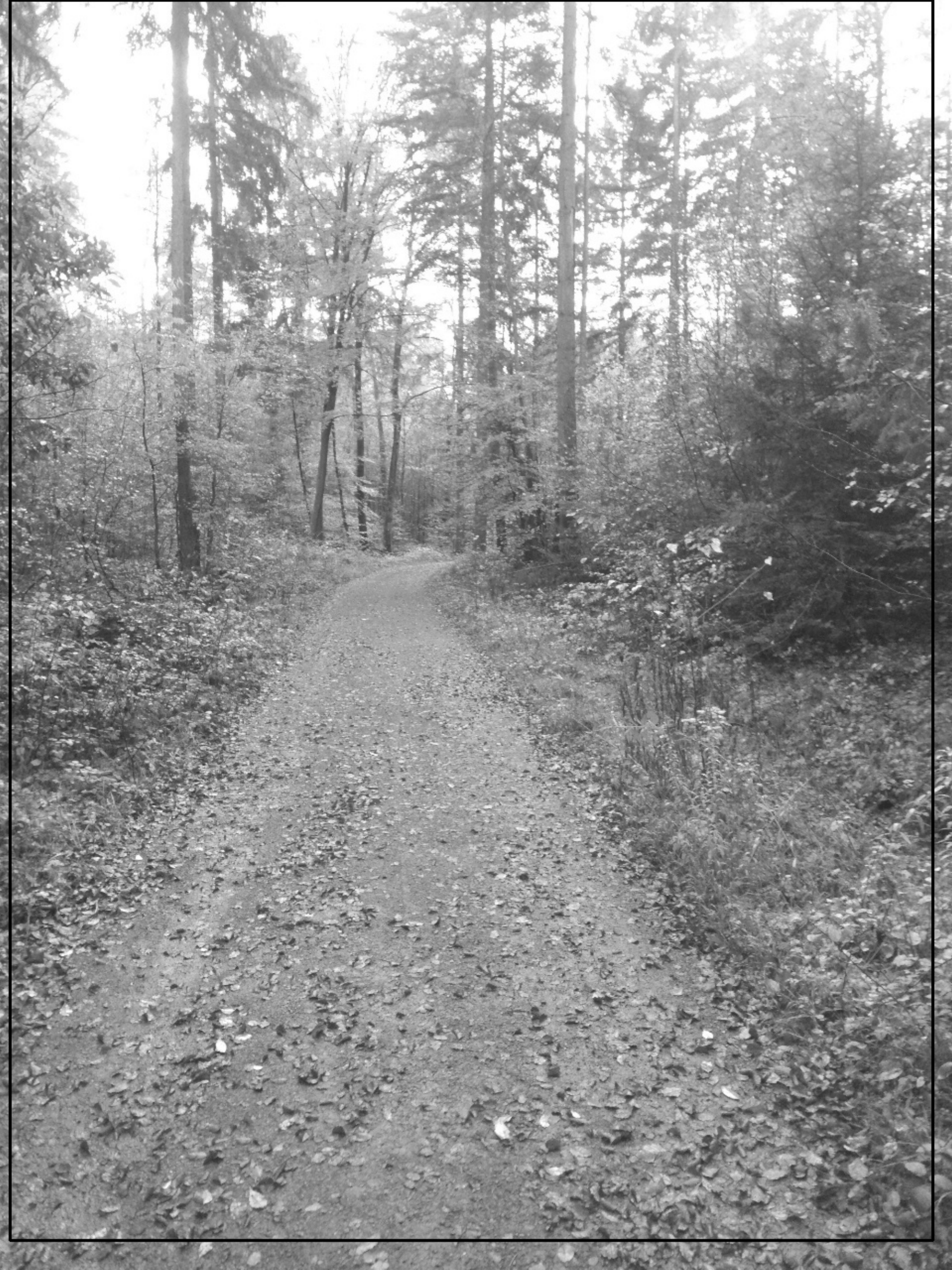} \tabularnewline
        \midrule
        \centering \includegraphics[angle=-180,origin=c, width=0.16\textwidth]{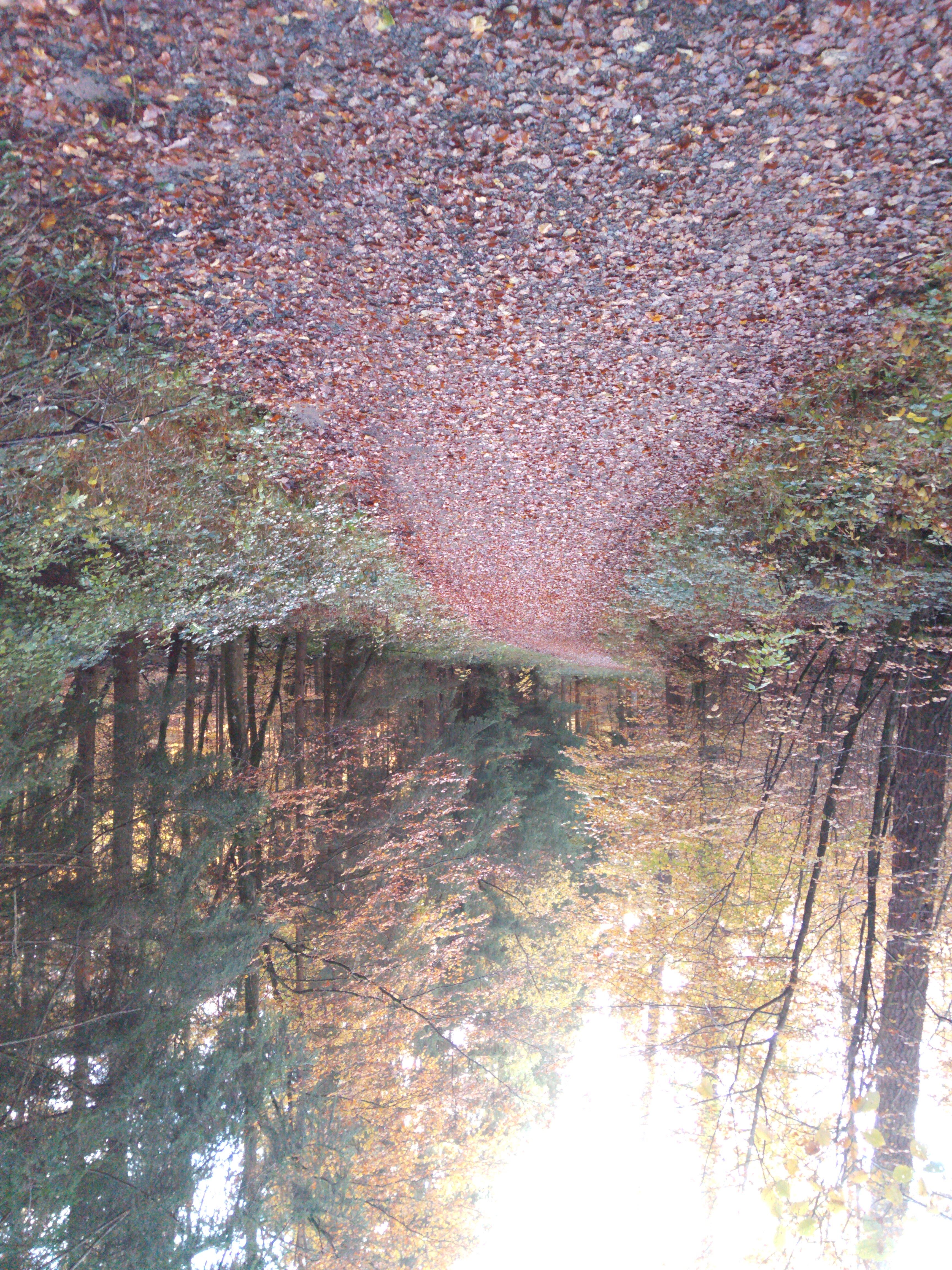} & 
        \centering \includegraphics[width=0.16\textwidth]{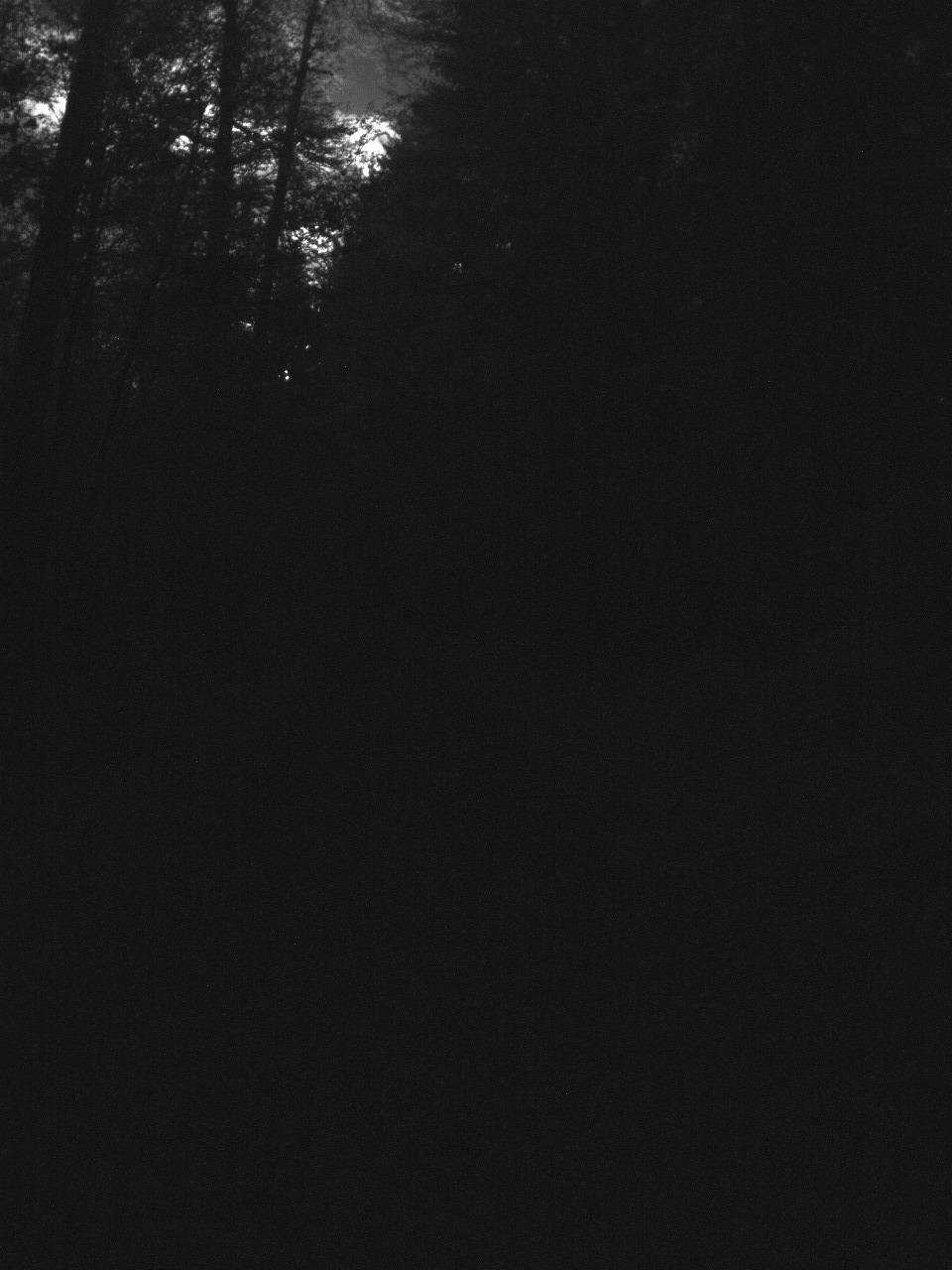} & 
        \centering \includegraphics[width=0.32\textwidth, height=0.22\textwidth]{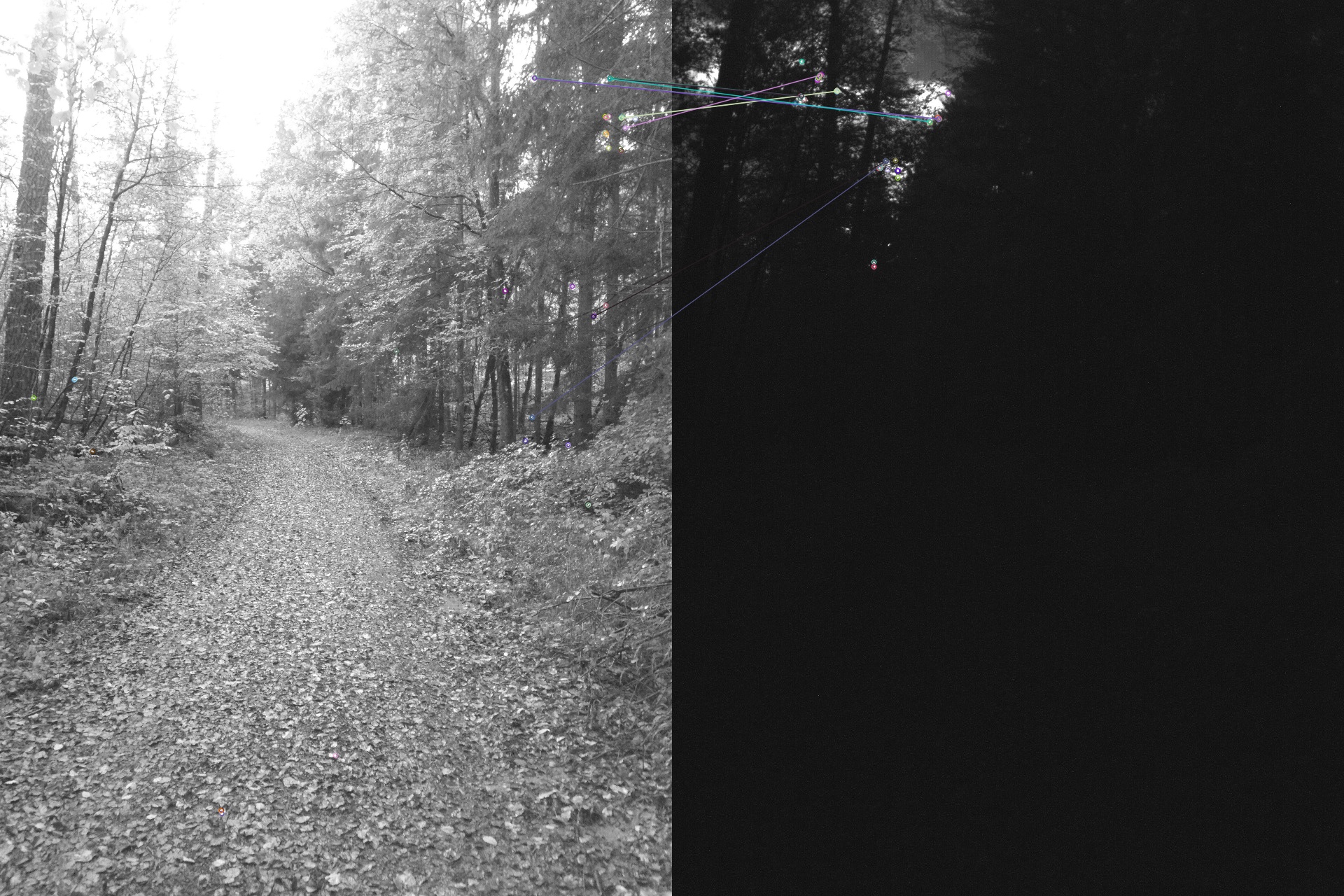} & 
        \centering \includegraphics[width=0.16\textwidth]{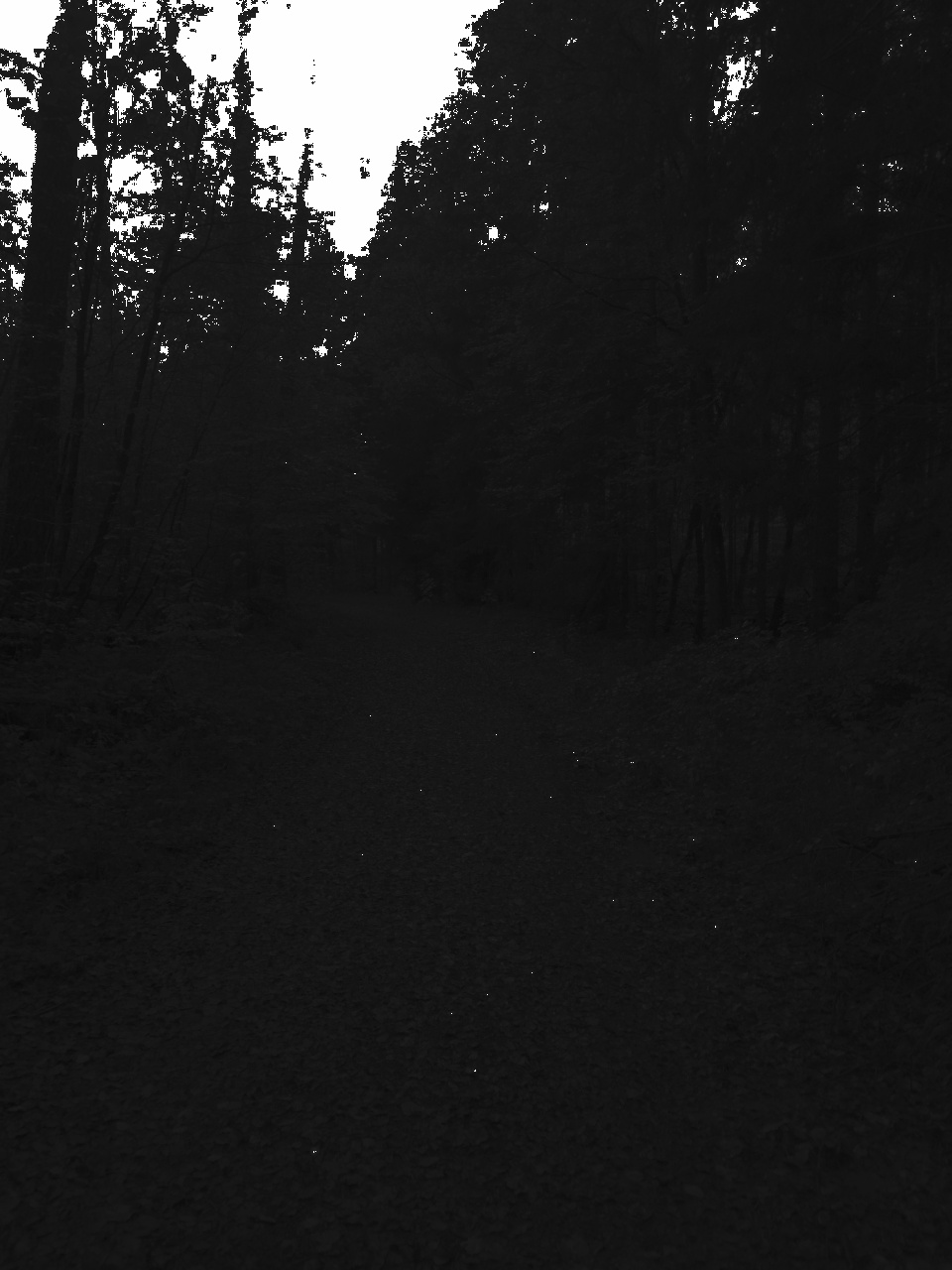} & 
        \centering \includegraphics[width=0.16\textwidth]{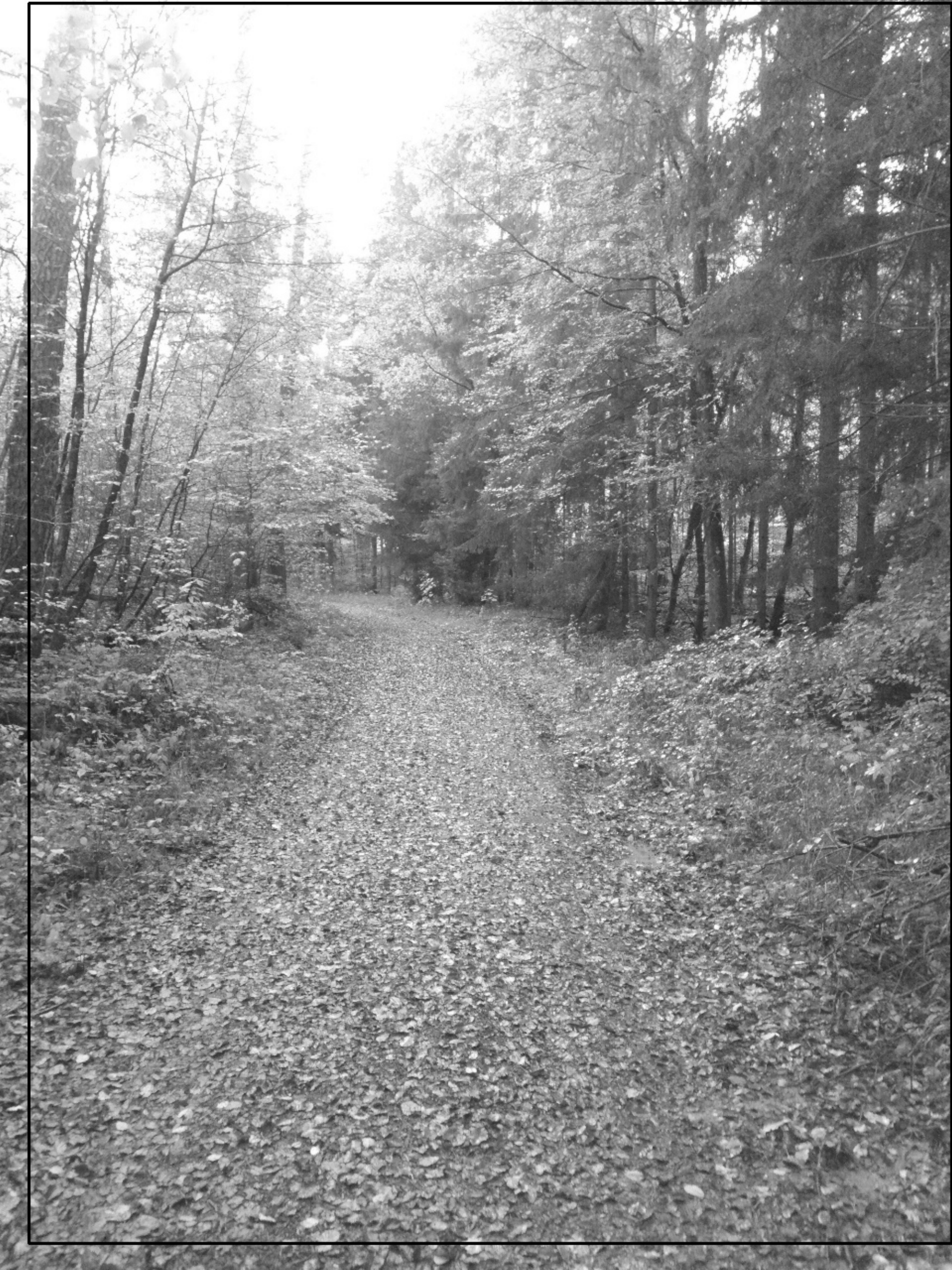} \tabularnewline
        \bottomrule
    \end{tabular}
    } % End resizebox

\end{table}

\begin{table}[htbp]
    \centering
    \renewcommand{\arraystretch}{1.3} % Adjust row spacing
    \setlength{\tabcolsep}{2pt} % Reduce column spacing for better compactness

    \caption{Comparison of Different feature matching Methods}
    \label{tab:traditional_methods}

    \resizebox{\columnwidth}{!}{ % Resize table to fit within the column width
    \begin{tabular}{m{0.18\textwidth} m{0.18\textwidth} m{0.22\textwidth} m{0.22\textwidth} m{0.22\textwidth}} 
        \toprule
        \centering \textbf{RGB Image} & 
        \centering \textbf{NIR Image} & 
        \centering \textbf{SIFT Matching} & 
        \centering \textbf{AKAZE Matching} & 
        \centering \textbf{ORB Matching} \tabularnewline
        \midrule
        % First Image Pair
        \centering \includegraphics[width=0.16\textwidth, height=0.16\textwidth]{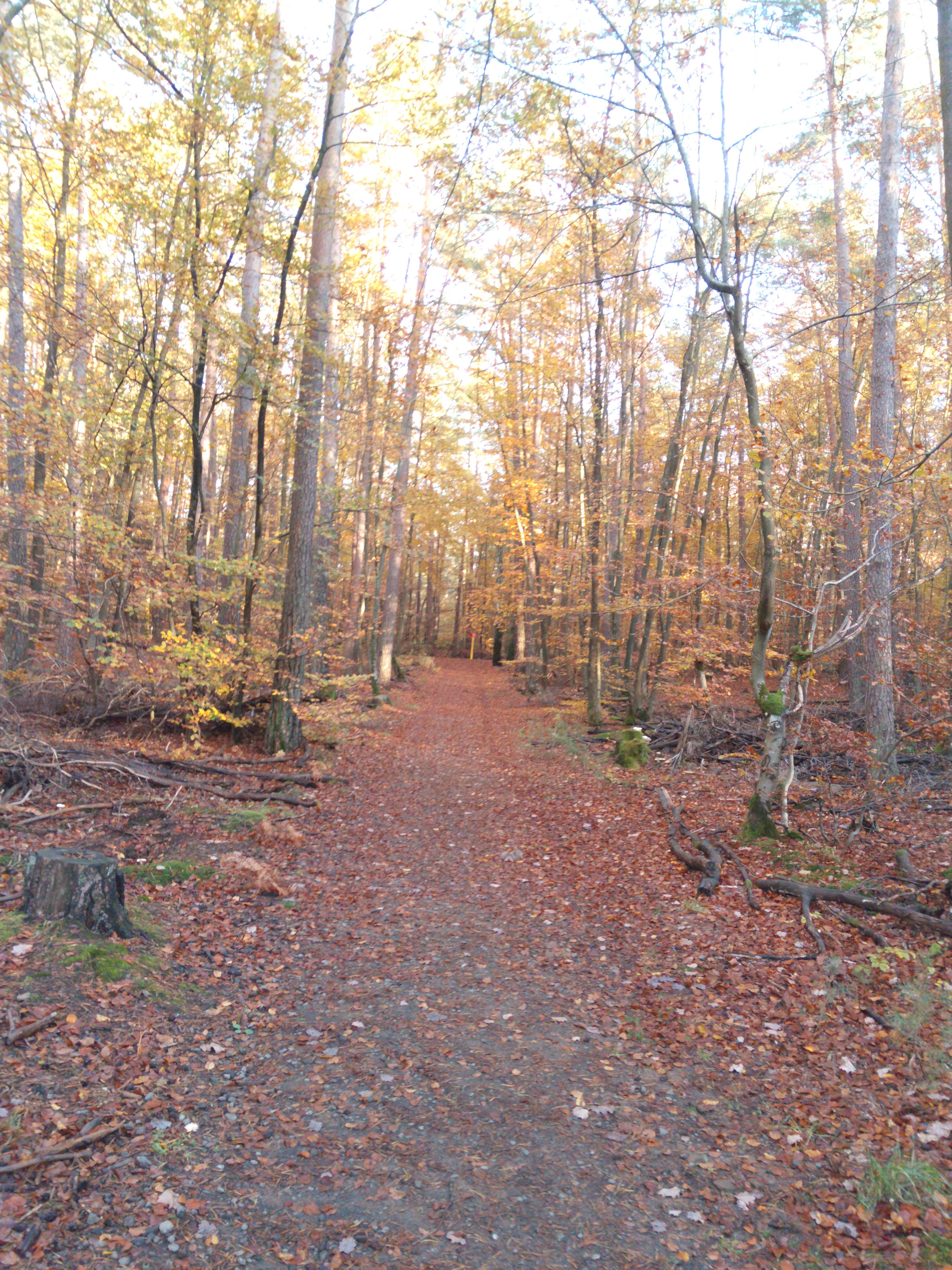} & 
        \centering \includegraphics[width=0.16\textwidth, height=0.16\textwidth]{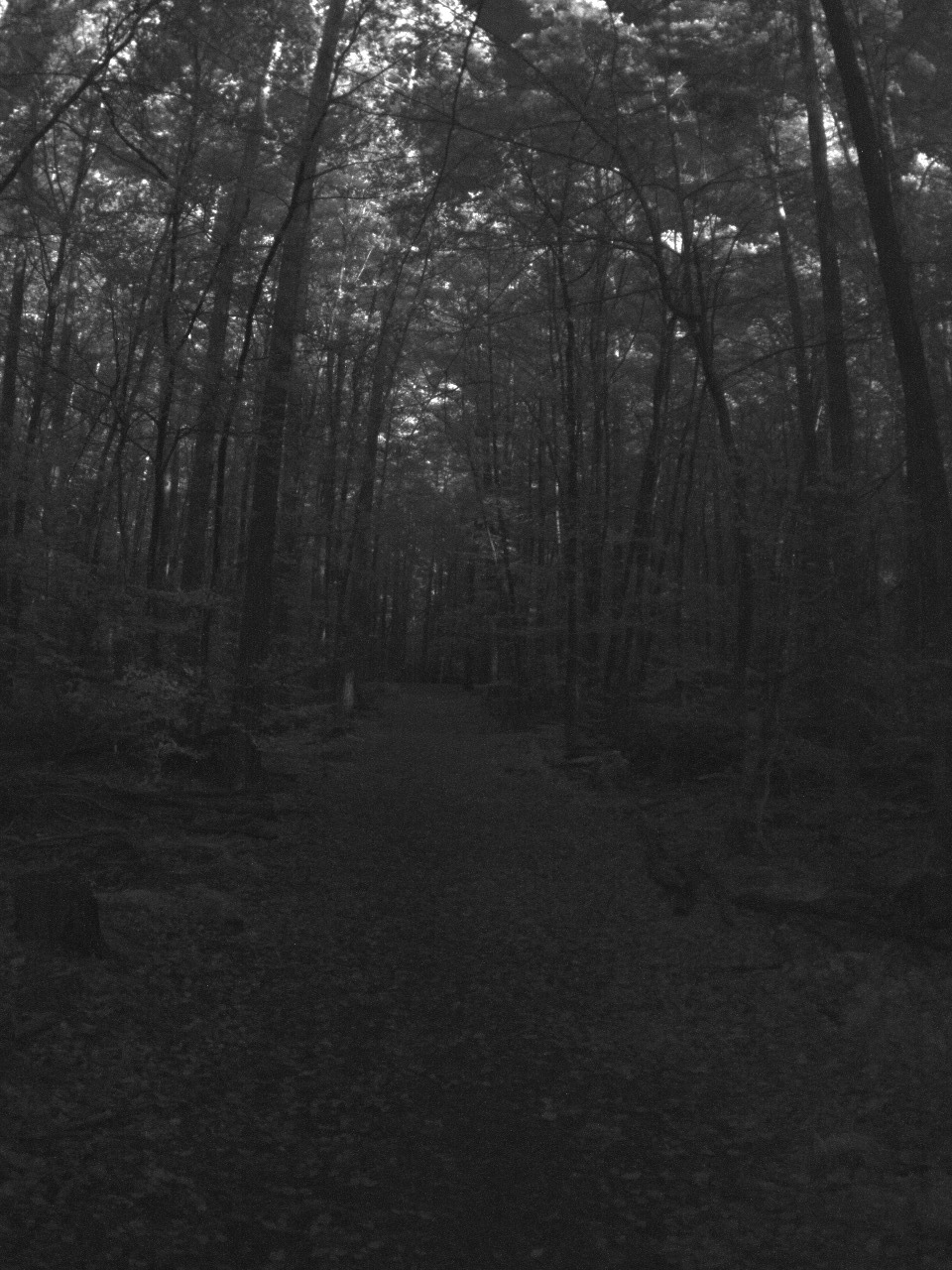} & 
        \centering \includegraphics[width=0.22\textwidth, height=0.16\textwidth]{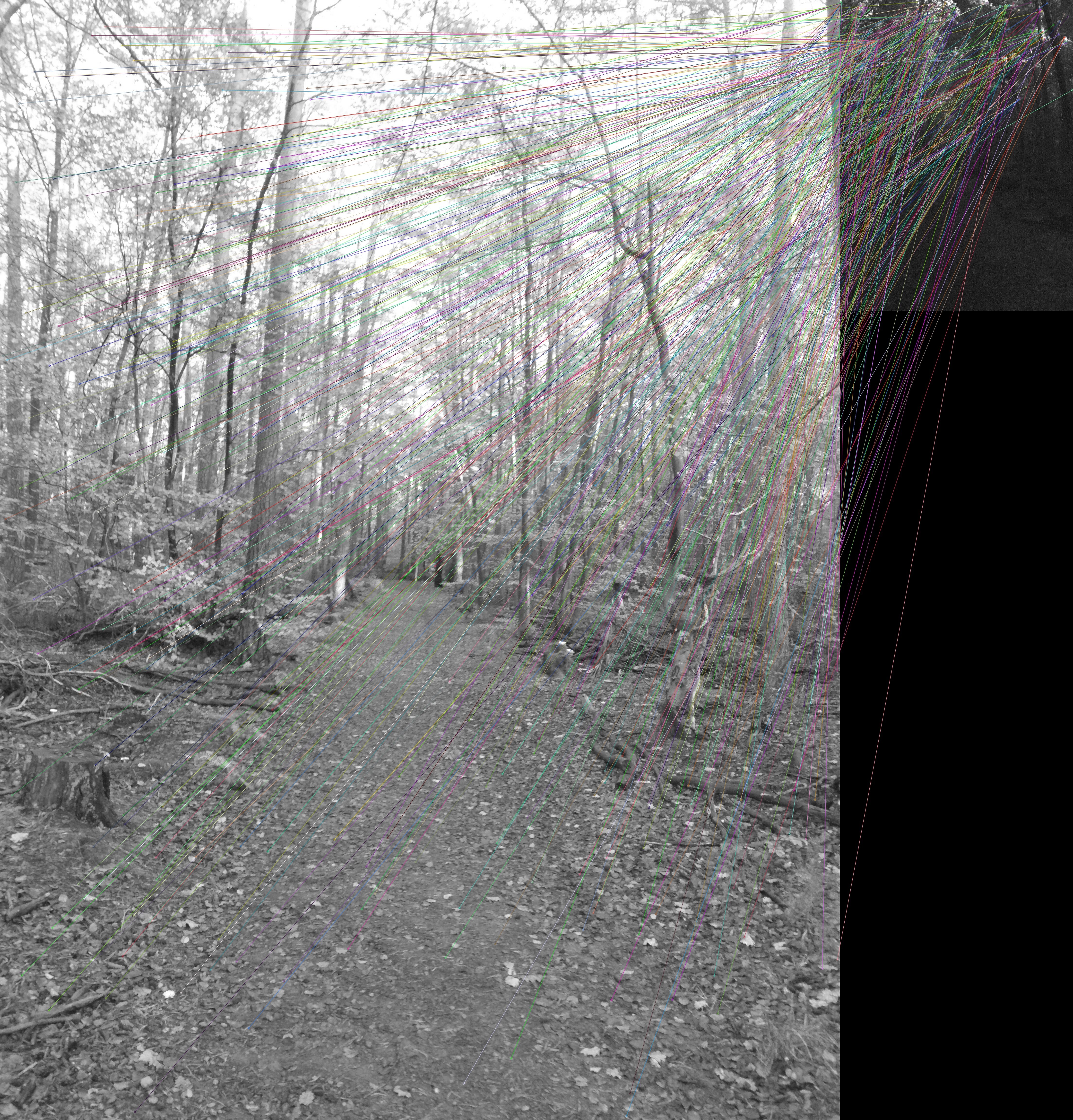} & 
        \centering \includegraphics[width=0.22\textwidth, height=0.16\textwidth]{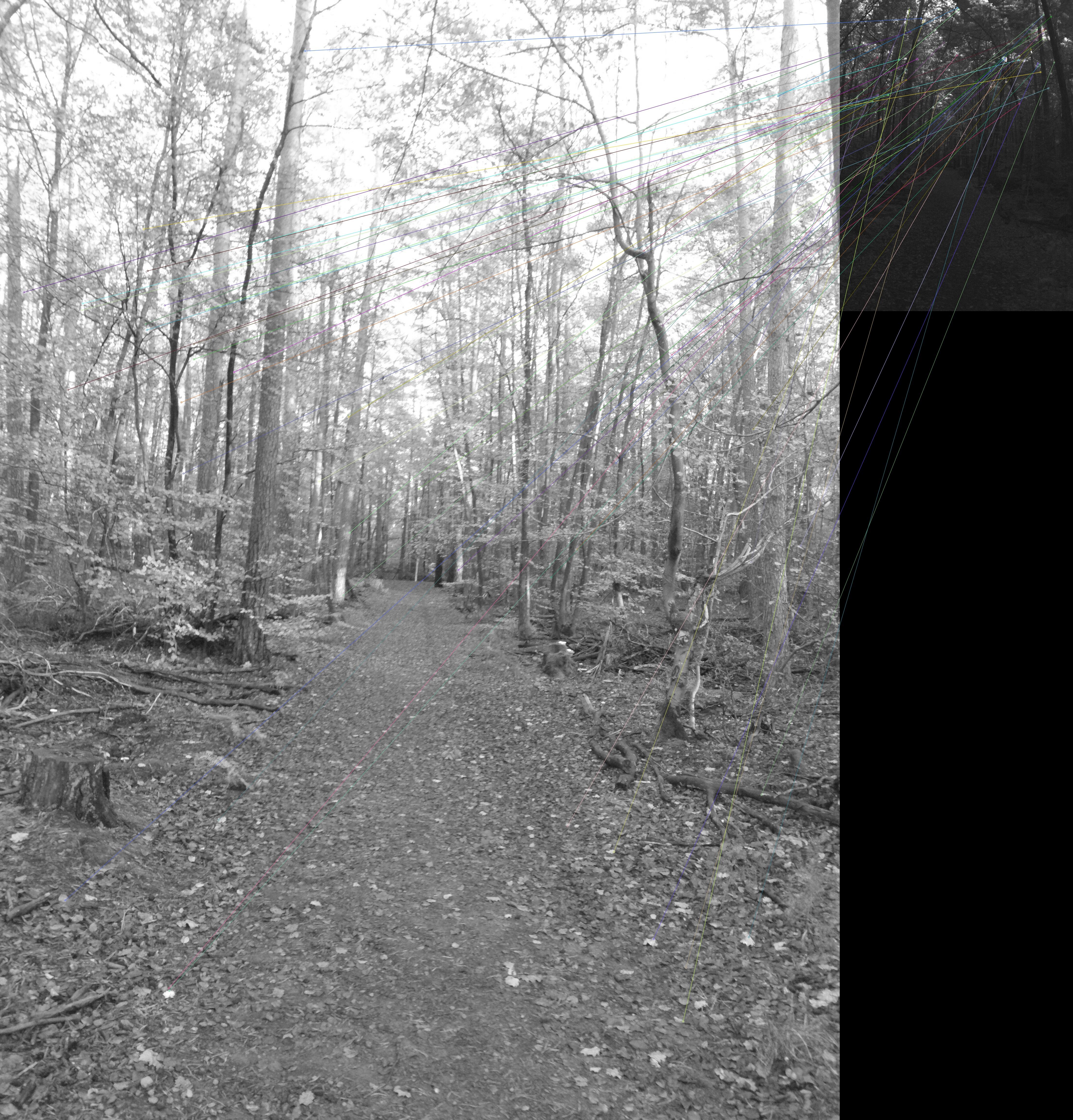} & 
        \centering \includegraphics[width=0.22\textwidth, height=0.16\textwidth]{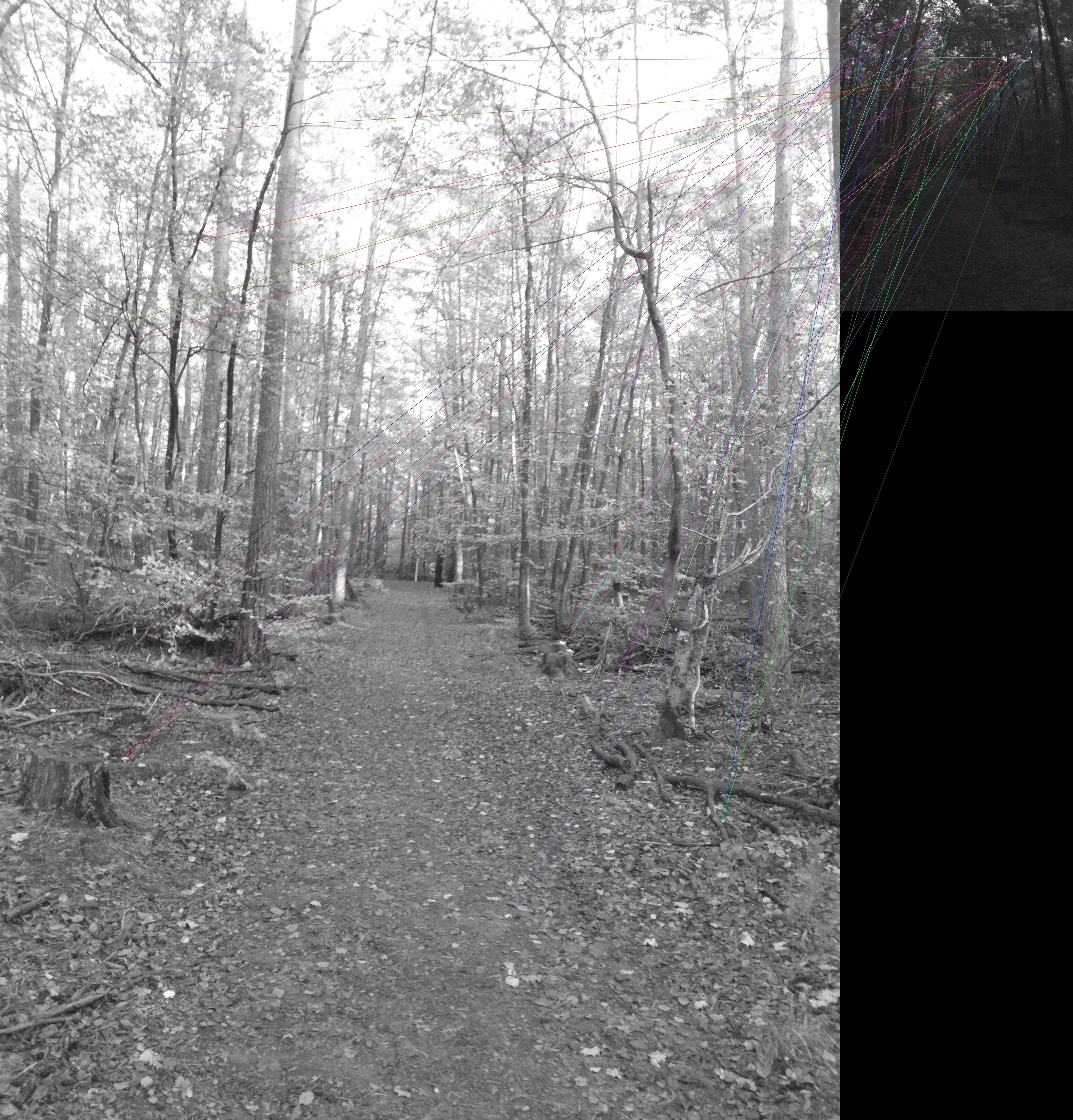} \tabularnewline
        \midrule
        % Second Image Pair
        \centering \includegraphics[angle=-180,origin=c, width=0.16\textwidth, height=0.16\textwidth]{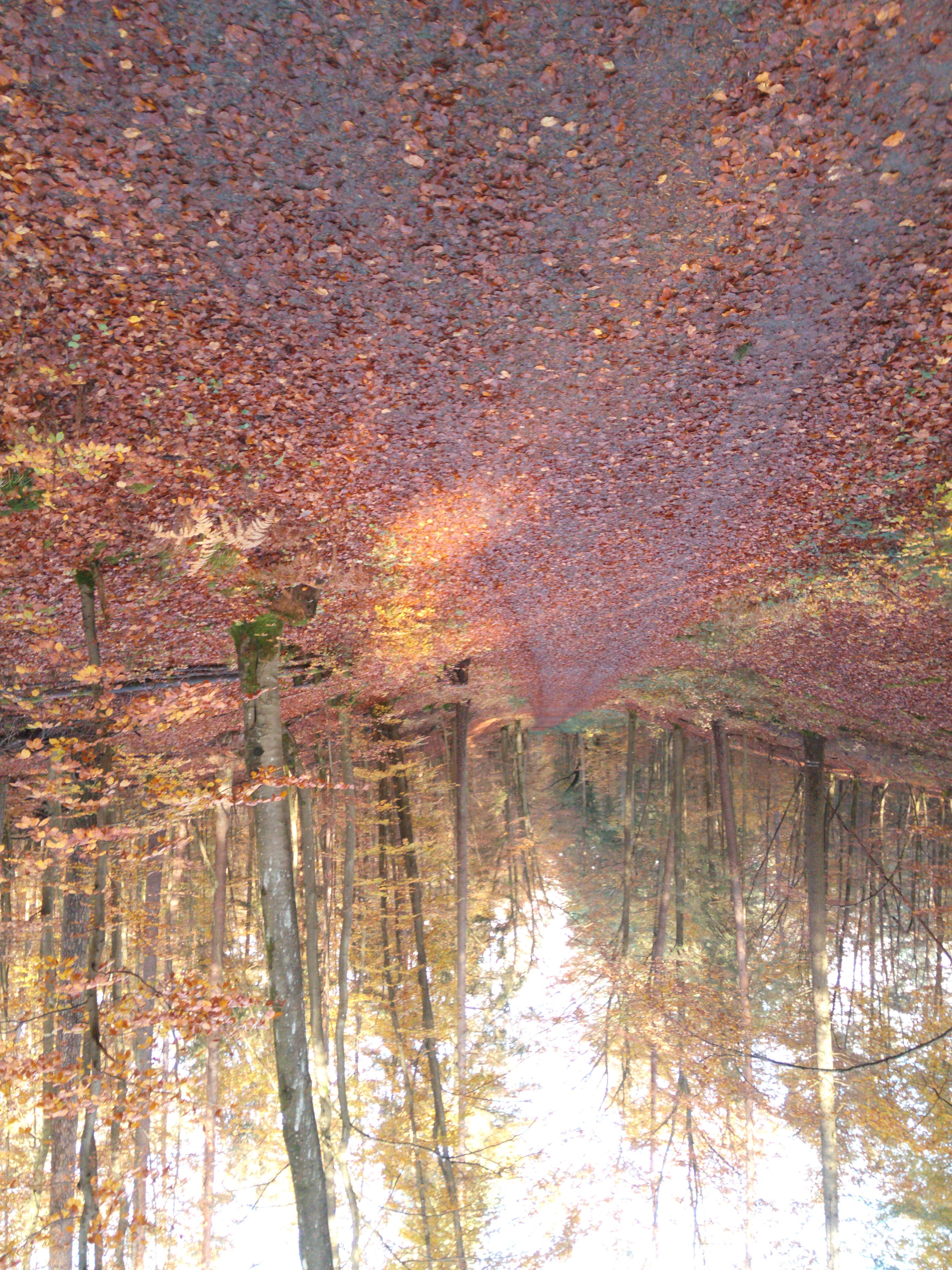} & 
        \centering \includegraphics[width=0.16\textwidth, height=0.16\textwidth]{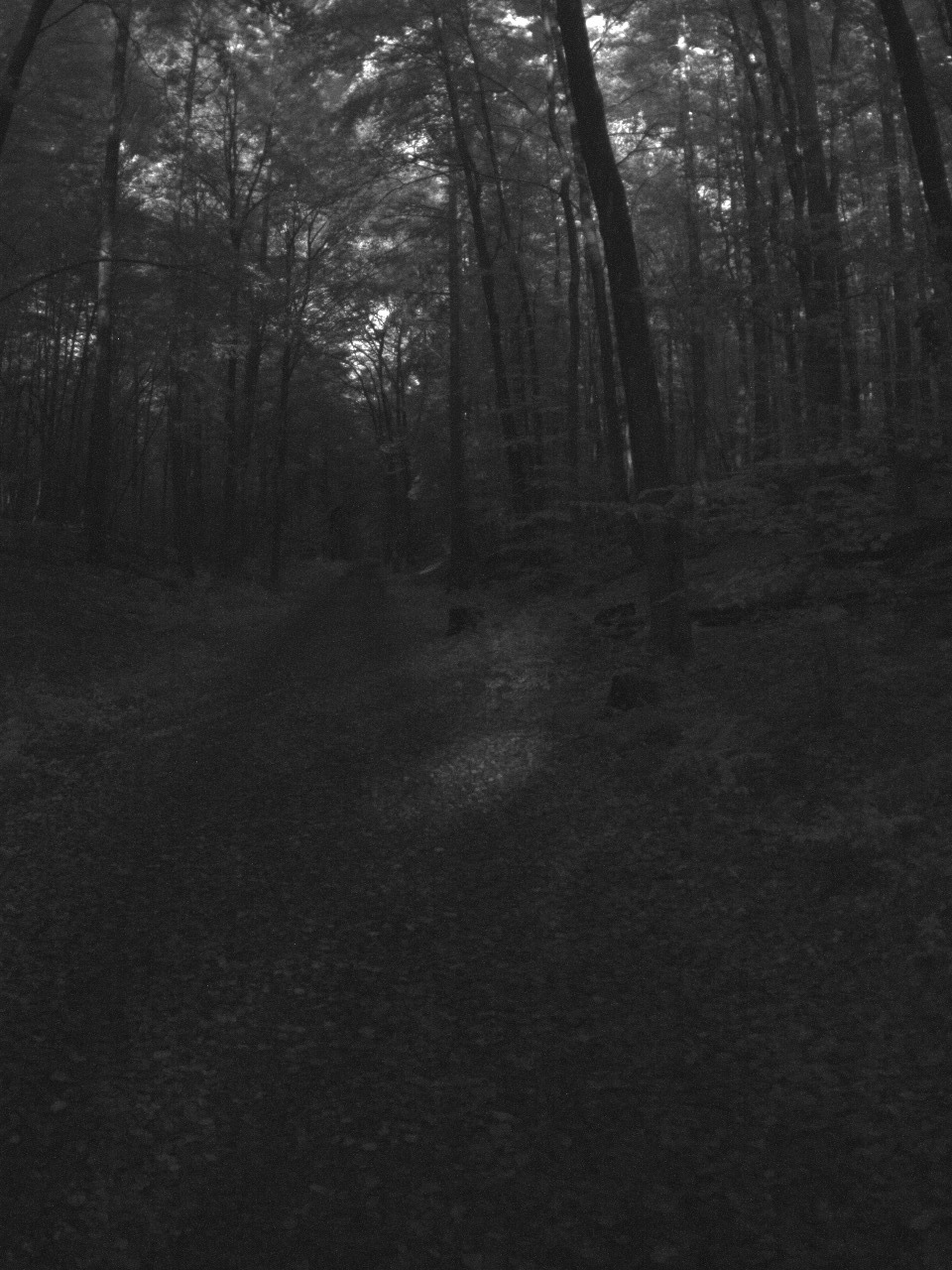} & 
        \centering \includegraphics[width=0.22\textwidth, height=0.16\textwidth]{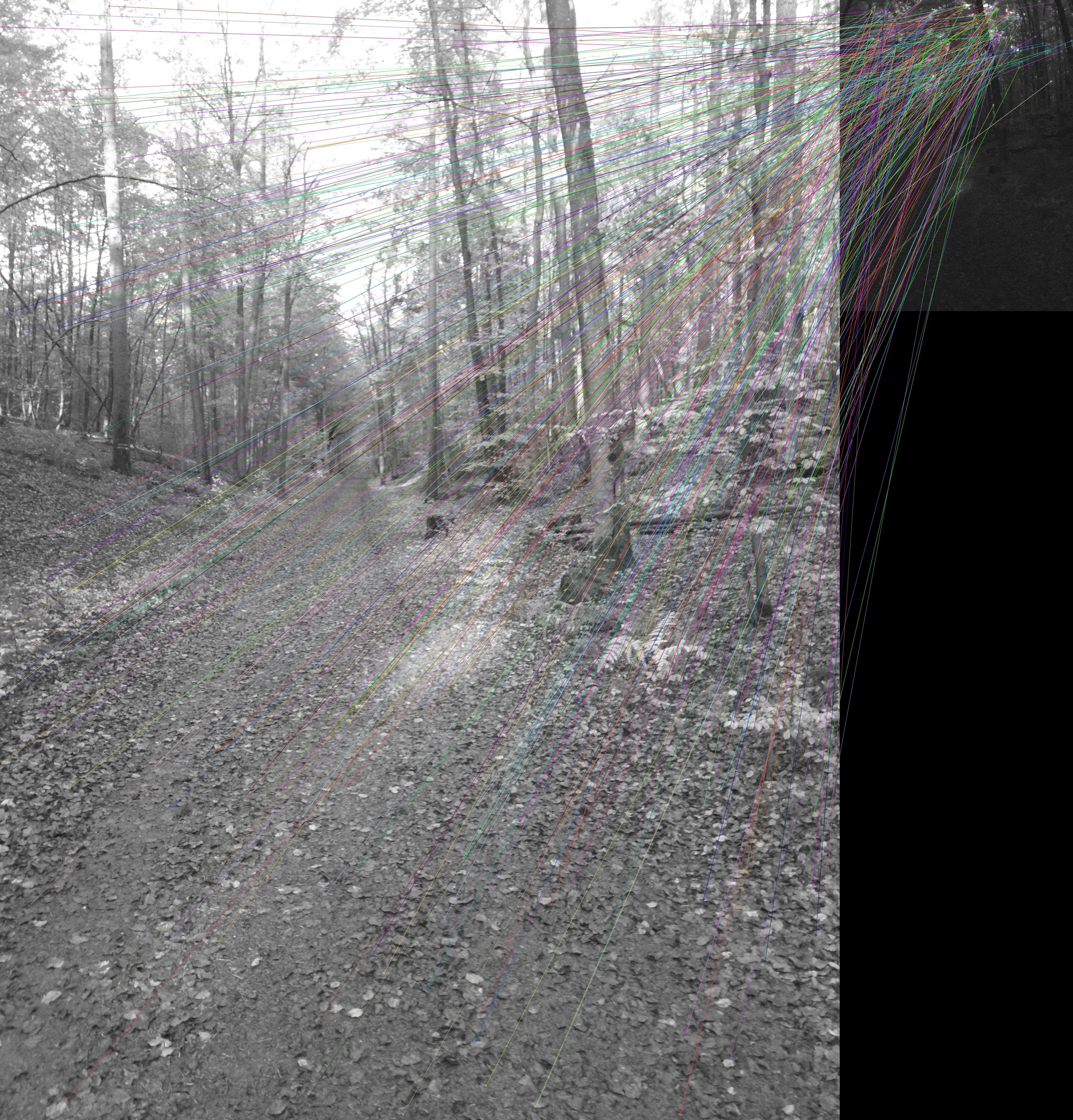} & 
        \centering \includegraphics[width=0.22\textwidth, height=0.16\textwidth]{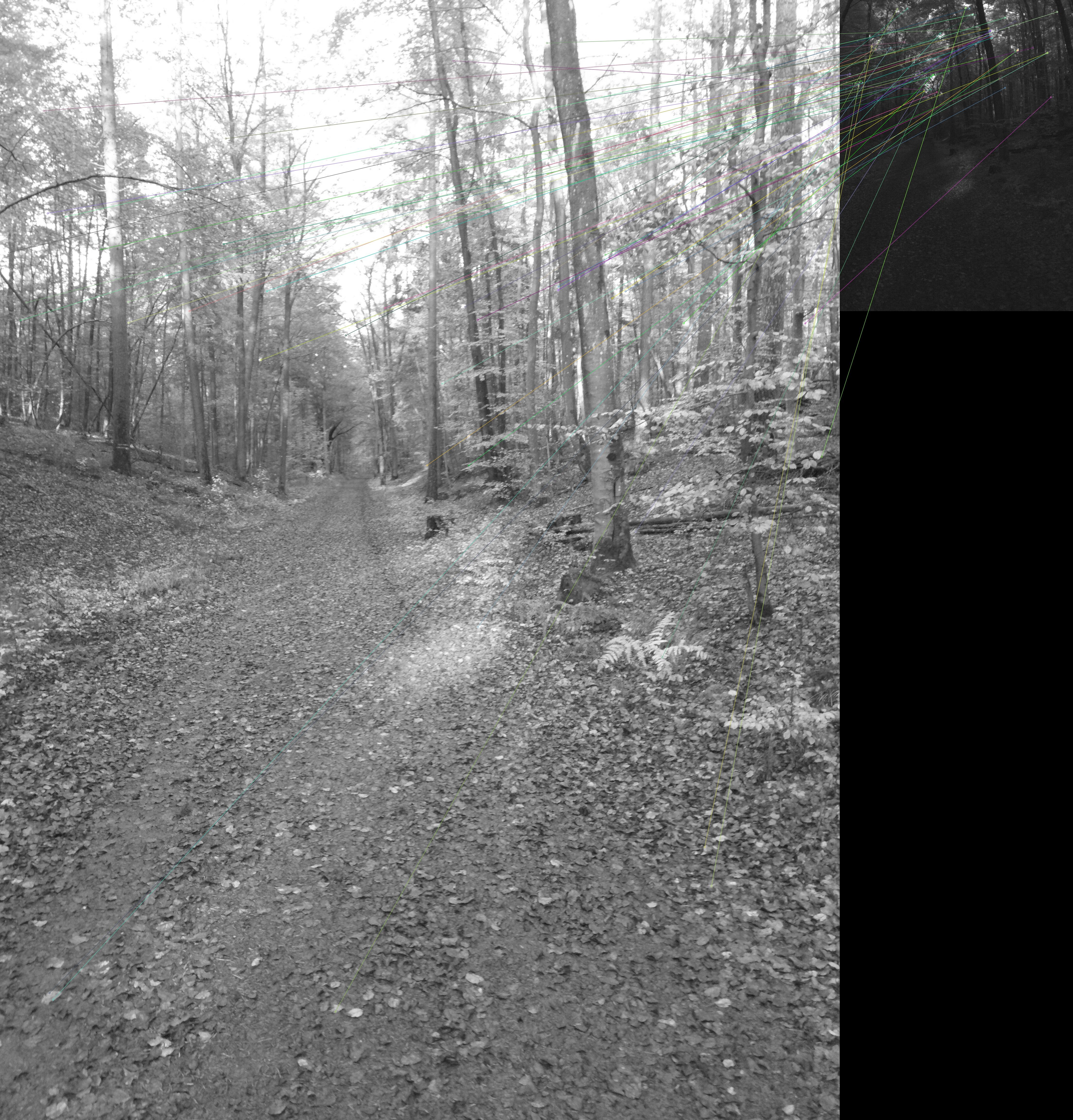} & 
        \centering \includegraphics[width=0.22\textwidth, height=0.16\textwidth]{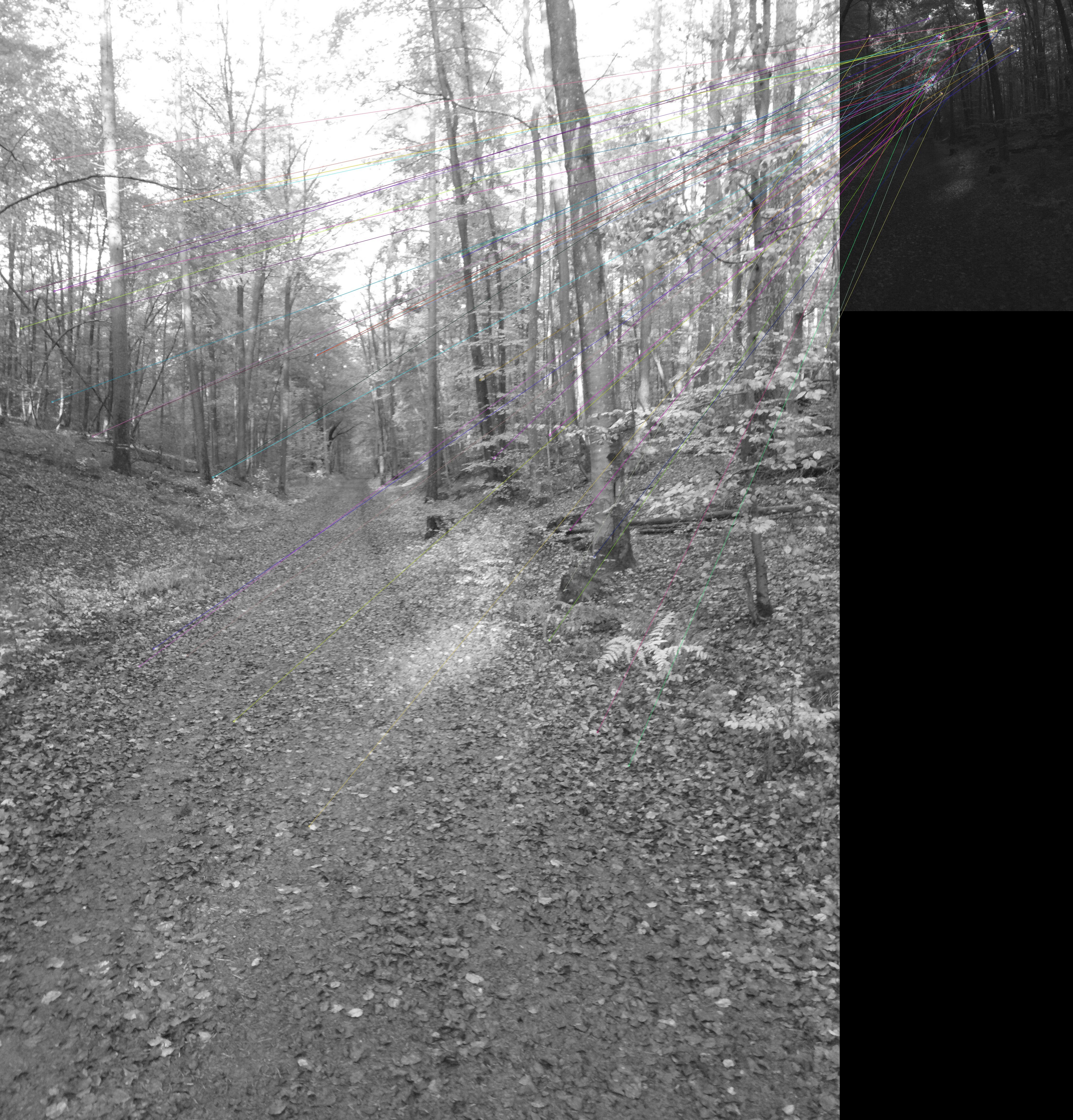} \tabularnewline
        \bottomrule
    \end{tabular}
    } % End resizebox

\end{table}

The results of feature matching between RGB/NIR images using SIFT, AKAZE and ORB are presented in table \ref{tab:traditional_methods} the analysis of these methods highlights significant challenges in cross-modal feature correspondence. SIFT, despite detecting the highest no. of keypoints, showcased a large no. of incorrect matches due to spectral differences. AKAZE performed somewhat better by reducing false correspondences, but still struggled with structural mismatches caused by intensity variation. ORB, being a binary descriptor-based method produced the smallest amount of matches, filtering out many incorrect correspondences, but at the cost of losing potentially useful matches. Overall, as these methods rely on intensity-based descriptors, the feature matching techniques are therefore non-robust.

\end{document}